\documentclass{article}

\usepackage{docmute}

\usepackage{arxiv}

\usepackage{graphicx}
\graphicspath{{fig/}{p3net-outputs/fig/}}

\usepackage[caption=false,font=footnotesize,labelfont=sf,textfont=sf]{subfig}

\usepackage[utf8]{inputenc}
\usepackage[T1]{fontenc}

\usepackage{amsmath}
\usepackage{amssymb}
\usepackage{amsfonts}
\usepackage{amsthm}
\usepackage{bm}
\usepackage{comment}
\usepackage{fancybox}
\usepackage{framed}
\usepackage{color}
\usepackage[dvipsnames]{xcolor}
\usepackage{multicol}
\usepackage{multirow}
\usepackage[para,online,flushleft]{threeparttable}
\usepackage{hyperref}
\usepackage{url}

\usepackage{enumitem}

\usepackage{varwidth}

\hypersetup{
  colorlinks = true,
  urlcolor = blue,
  linkcolor = red,
  citecolor = green
}

\usepackage{algorithm}
\usepackage[noend]{algpseudocode}
\usepackage{algorithmicx}

\expandafter\def\expandafter\UrlBreaks\expandafter{\UrlBreaks
  \do\a\do\b\do\c\do\d\do\e\do\f\do\g\do\h\do\i\do\j%
  \do\k\do\l\do\m\do\n\do\o\do\p\do\q\do\r\do\s\do\t%
  \do\u\do\v\do\w\do\x\do\y\do\z\do\A\do\B\do\C\do\D%
  \do\E\do\F\do\G\do\H\do\I\do\J\do\K\do\L\do\M\do\N%
  \do\O\do\P\do\Q\do\R\do\S\do\T\do\U\do\V\do\W\do\X%
  \do\Y\do\Z}

\algrenewcommand\algorithmicthen{:}

\algnewcommand\algorithmicforeach{\textbf{for each}}
\algdef{S}[FOR]{ForEach}[1]{\algorithmicforeach\ #1\ \algorithmicdo}

\algnewcommand\AlgAnd{\textbf{and} }
\algnewcommand\AlgOr{\textbf{or} }
\algnewcommand\AlgBreak{\textbf{break}}
\algnewcommand\AlgContinue{\textbf{continue}}

\algrenewcommand\textproc{}

\algrenewcommand\alglinenumber[1]{\color{black}\footnotesize #1:}

\algnewcommand{\Initialize}[1]{
  \State \textbf{Initialize:}
  \State \hspace*{\algorithmicindent}\parbox[t]{0.8\linewidth}{\raggedright #1}}

\algnewcommand{\LeftComment}[1]{
  \Statex $\triangleright$ #1 \hfill}

\algnewcommand{\IIf}[1]{\State\algorithmicif\ #1\ \algorithmicthen}
\algnewcommand{\EndIIf}{\unskip}

\def\BibTeX{{\rm B\kern-.05em{\sc i\kern-.025em b}\kern-.08em
  T\kern-.1667em\lower.7ex\hbox{E}\kern-.125emX}}

\newcommand{\ProposalName}{P3Net}
\newcommand{\ProposalNameFPGA}{P3NetFPGA}
\newcommand{\IPName}{P3NetCore}
\newcommand{\NewENet}{ENetLite}
\newcommand{\NewPNet}{PNetLite}
\newcommand{\EPNet}{\{E, P\}Net}
\newcommand{\NewEPNet}{\{E, P\}NetLite}

\title{An Integrated FPGA Accelerator for \\ Deep Learning-based 2D/3D Path Planning}

\renewcommand{\shorttitle}{An Integrated FPGA Accelerator for Deep Learning-based 2D/3D Path Planning}

\author{
  \href{https://orcid.org/0000-0001-8534-2381}{
    \includegraphics[scale=0.075]{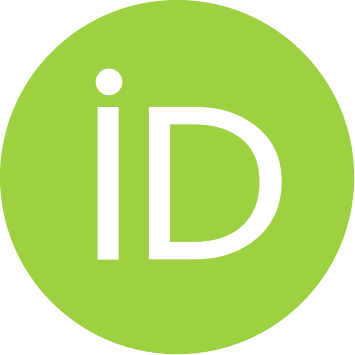}}\hspace{1mm}Keisuke Sugiura \\
  Keio University \\
  3-14-1 Hiyoshi, Kohoku-ku, Yokohama, Japan \\
  \texttt{sugiura@arc.ics.keio.ac.jp} \\
  \And
  \href{https://orcid.org/0000-0001-9578-3842}{
    \includegraphics[scale=0.075]{orcid.pdf}}\hspace{1mm}Hiroki Matsutani \\
  Keio University \\
  3-14-1 Hiyoshi, Kohoku-ku, Yokohama, Japan \\
  \texttt{matutani@arc.ics.keio.ac.jp}
}

\hypersetup{
  pdftitle={An Integrated FPGA Accelerator for Deep Learning-based 2D/3D Path Planning},
  pdfsubject={cs.AR},
  pdfauthor={Keisuke Sugiura, Hiroki Matsutani},
  pdfkeywords={Path planning, Neural path planning, Point cloud processing, PointNet, Deep learning, FPGA}
}

\begin{document}


\maketitle



\begin{abstract}
Path planning is a crucial component for realizing the autonomy of mobile robots.
However, due to limited computational resources on mobile robots, it remains challenging to deploy state-of-the-art methods and achieve real-time performance.
To address this, we propose {\ProposalName} (\underline{P}ointNet-based \underline{P}ath \underline{P}lanning \underline{Net}works), a lightweight deep-learning-based method for 2D/3D path planning, and design an IP core ({\IPName}) targeting FPGA SoCs (Xilinx ZCU104).
{\ProposalName} improves the algorithm and model architecture of the recently-proposed MPNet.
{\ProposalName} employs an encoder with a PointNet backbone and a lightweight planning network in order to extract robust point cloud features and sample path points from a promising region.
{\IPName} is comprised of the fully-pipelined point cloud encoder, batched bidirectional path planner, and parallel collision checker, to cover most part of the algorithm.
On the 2D (3D) datasets, {\ProposalName} with the IP core runs 24.54--149.57x and 6.19--115.25x (10.03--59.47x and 3.38--28.76x) faster than ARM Cortex CPU and Nvidia Jetson while only consuming 0.255W (0.809W), and is up to 1049.42x (133.84x) power-efficient than the workstation.
{\ProposalName} improves the success rate by up to 28.2\% and plans a near-optimal path, leading to a significantly better tradeoff between computation and solution quality than MPNet and the state-of-the-art sampling-based methods.


\end{abstract}

\section{Introduction} \label{sec:intro}


Path planning aims to find a feasible path from a start to a goal position while avoiding obstacles.
It is a fundamental component for mobile robots to autonomously navigate and accomplish a variety of tasks, e.g., farm monitoring~\cite{JeonghyeonPak22}, aerial package delivery~\cite{HyeonbeomLee18}, mine exploration~\cite{TungDang19}, and rescue in a collapsed building~\cite{FrancisColas13}.
Such robotic applications are often deployed on resource-limited edge devices due to severe constraints on the cost and payload.
In addition, real-time performance is of crucial importance, since robots may have to plan and update a path on-the-fly in dynamic environments, and delays in path planning would affect the stability of the upstream applications.
To cope with the strict performance requirements, FPGA SoCs are increasingly used in robotic applications such as visual odometry~\cite{RunzeLiu19} and SLAM~\cite{PengfeiGu22}.
FPGA SoC integrates an embedded CPU with a reconfigurable fabric, which allows to develop a custom accelerator tailored for a specific algorithm.
Taking these into account, an FPGA-based efficient path planning implementation becomes an attractive solution, which would greatly broaden the application range, since mobile robots can now perform expensive planning tasks on its own without connectivity to remote servers.

\begin{figure}[t!]
  \begin{tabular}{c}
    \begin{minipage}[t]{\linewidth}
      \centering
      \includegraphics[keepaspectratio, width=0.7\linewidth]{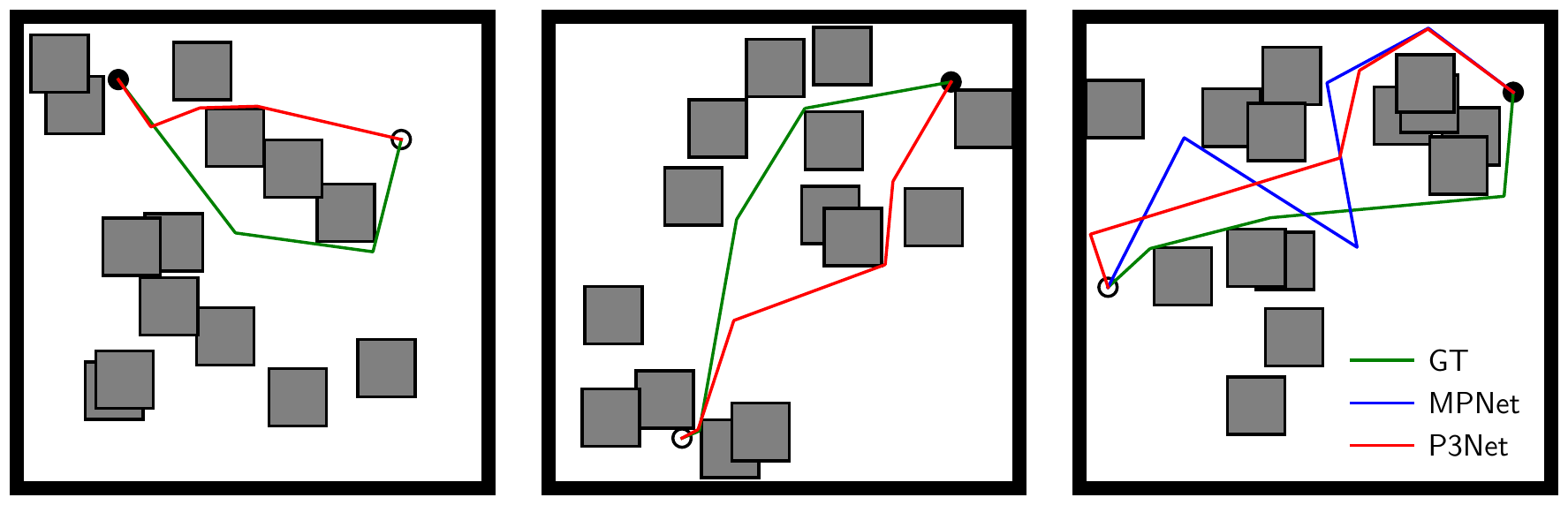}
      \caption{Results on the {\ProposalName}2D dataset.
      While MPNet (\textcolor{blue}{blue}) fails to plan feasible paths in the first two cases, the proposed {\ProposalName} (\textcolor{red}{red}) plans successfully in all these cases, while reducing the parameters by 32.32x.}
      \label{fig:result-p3net2d}
    \end{minipage} \vspace*{10pt} \\
    \begin{minipage}[t]{\linewidth}
      \centering
      \includegraphics[keepaspectratio, width=0.7\linewidth]{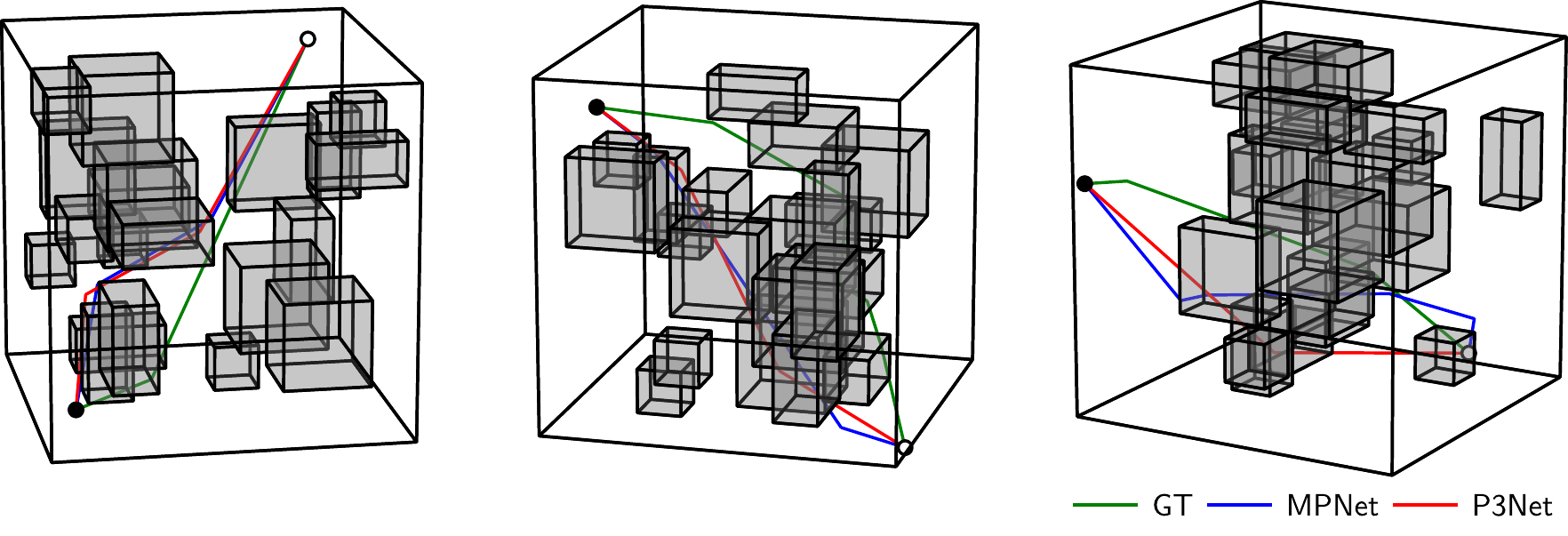}
      \caption{Results on the {\ProposalName}3D dataset.
      {\ProposalName} (\textcolor{red}{red}) produces feasible paths in these cases with 5.43x less parameters than MPNet.}
      \label{fig:result-p3net3d}
    \end{minipage}
  \end{tabular}
\end{figure}

In path planning, the sampling-based methods including Rapidly-exploring Random Tree (RRT)~\cite{StevenLaValle98} and RRT*~\cite{SertacKaraman11} are the most prominent, owing to their simplicity and theoretical guarantees.
Instead of working on a discretized grid environment like graph-based methods (e.g., A*~\cite{PeterHart68}), these methods explore the environment by incrementally building a tree that represents a set of valid robot motions.
While this alleviates the curse of dimensionality to some extent, the computational and memory cost for finding a near-optimal solution is still high, since a sufficient number of tree nodes should be placed to fill the entire free space.
A number of RRT variants, e.g., Informed-RRT*~\cite{JonathamGammell14} and BIT*~\cite{JonathamGammell15}, have been proposed to improve the sampling efficiency and convergence speed.
While they offer better tradeoffs between computational effort and solution quality, they rely on carefully designed heuristics, which imply the prior knowledge of the environment and may not be effective in certain scenarios.
Due to their increased algorithmic complexity, it even takes up to tens of seconds to complete a task on an embedded CPU.
On top of that, their inherently sequential nature would require intricate strategies to map onto a parallel computing platform.


Motivated by the tremendous success of deep learning, the research effort is devoted to developing learning-based methods; the basic idea is to automatically acquire a policy for planning near-optimal paths from a large collection of paths generated by the existing methods.
MPNet~\cite{AhmedQureshi21} is a such recently-proposed method.
It employs two separate MLP networks for encoding and planning; the former embeds a point cloud representing obstacles into a latent feature space, while the latter predicts a next robot position to incrementally expand a path.
Unlike sampling-based methods, MPNet does not involve operations on complex data structures (e.g., KNN search on a K-d tree), and DNN inference is more amenable to parallel processing.
This greatly eases the design of a custom processor and makes MPNet a promising candidate for the low-cost FPGA implementation.
Its performance is however limited due to the following reasons; the encoder does not take into account the unstructured and unordered nature of point clouds, which degrades the quality of extracted features and eventually results in a lower success rate.
Furthermore, the planning network has a lower parameter efficiency, and MPNet has a limited parallelism as it processes only one candidate path at a time until a feasible solution is found.


This paper addresses the above limitations of MPNet and proposes a new learning-based method for 2D/3D path planning, named \textbf{{\ProposalName}} (\underline{P}ointNet-based \underline{P}ath \underline{P}lanning \underline{Net}works), along with its custom IP core for FPGAs (\textbf{{\IPName}}).
While the existing methods often assume the availability of abundant computing resources (e.g., GPUs), which is not the case in practice, {\ProposalName} is designed to work on resource-limited edge devices and still deliver satisfactory performance.
Besides, to our knowledge, {\IPName} is one of the first FPGA-based accelerators for fully learning-based path planning.
The main contributions of this paper are summarized as follows:
\begin{enumerate}[leftmargin=*]
  \item To extract robust features and improve parameter efficiency, we utilize a PointNet~\cite{CharlesQi17}-based encoder architecture, which is specifically designed for point cloud processing, together with a lightweight planning network.
  PointNet is widely adopted as a backbone for point cloud tasks including classification~\cite{XuMa21} and segmentation~\cite{WeiyueWang18}.

  \item We make two algorithmic modifications to MPNet; we introduce a batch planning strategy to process multiple candidate paths in parallel, which offers a new parallelism and improves the success rates without increasing the computation time.
We then add a refinement phase at the end to iteratively optimize the path.

  \item We design a custom IP core for {\ProposalName}, which integrates a fully-pipelined point cloud encoder, a bidirectional neural planner, and a parallelized collision checker.
{\ProposalName} is developed using High-Level Synthesis (HLS) and implemented on a Xilinx ZCU104 Evaluation Kit.

  \item Experimental results validate the effectiveness of the proposed algorithmic optimizations and new models.
FPGA-accelerated {\ProposalName} achieves significantly better tradeoff between computation time and success rate than MPNet and the state-of-the-art sampling-based methods.
It runs up to two orders of magnitude faster than the embedded CPU (ARM Cortex) or integrated CPU-GPU systems (Nvidia Jetson).
{\ProposalName} quickly converges to near-optimal solutions in most cases, and is up to three orders of magnitude more power-efficient than the workstation.
\end{enumerate}
The rest of the paper is structured as follows: Section \ref{sec:related} provides a brief overview of the related works, while Section \ref{sec:prelim} covers the preliminaries.
The proposed {\ProposalName} is explained in Section \ref{sec:method}, and Section \ref{sec:impl} elaborates on the design and implementation of {\IPName}.
Experimental results are presented in Section \ref{sec:eval}, and Section \ref{sec:conc} concludes the paper.

\section{Related Works} \label{sec:related}

\subsection{Sampling and Learning-based Path Planning} \label{sec:related-learning-based}
The sampling-based methods, e.g., RRT~\cite{StevenLaValle98} and its provably asymptotically-optimal version, RRT*~\cite{SertacKaraman11}, are prevalent in robotic path planning; they explore the environment by incrementally growing an exploration tree.
Considering that the free space should be densely filled with tree nodes to find a high-quality solution, the computational complexity is at worst exponential with respect to the space dimension and may be intractable in some difficult cases such as the environments with narrow passages.
The later methods introduce various heuristics to improve search efficiency; Informed-RRT*~\cite{JonathamGammell14} uses an ellipsoidal heuristic, while BIT*~\cite{JonathamGammell15} and its variants~\cite{MarlinPStrub20A,MarlinPStrub20B} apply graph-search techniques.
Despite the steady improvement, they still rely on sophisticated heuristics; deep learning-based methods have been extensively studied to automatically learn effective policies for planning high-quality paths.


Several studies have investigated the hybrid approach, where deep learning techniques are incorporated into the classical planners.
Ichter \textit{et al.}~\cite{BrianIchter18,BrianIchter19} and Wang \textit{et al.}~\cite{JiankunWang20} extend RRT by generating informed samples from a learned latent space.
Neural A*~\cite{RyoYonetani21} is a differentiable version of A*, while WPN~\cite{AlexandruIosifToma21} uses LSTM to generate path waypoints and then A* to connect them.
Aside from the hybrid approach, an end-to-end approach aims to directly learn the behavior of classical planners via supervised learning; Inoue \textit{et al.}~\cite{MasayaInoue18} and Bency \textit{et al.}~\cite{MayurBency19} train LSTM networks on the RRT* and A*-generated paths, respectively.
CNN~\cite{MarkPfeiffer17}, PointNet~\cite{JinwookHuh21,AdamFishman22}, and Transformer~\cite{DevendraSinghChaplot21,JacobJohnson21,MohitShridhar22} are also employed to construct end-to-end models.


MPNet and our {\ProposalName} follow the end-to-end supervised approach, and perform path planning on a continuous domain by directly regressing coordinates of the path points.
{\ProposalName} is unique in that it puts more emphasis on the computational and resource efficiency and builds upon a lightweight MLP network, making it suitable for deployment on the low-end edge devices.
Besides, the environment is represented by a space-efficient sparse point cloud as in \cite{RobinStrudel20,JinwookHuh21,AdamFishman22}, unlike the grid map which introduces quantization errors and leads to a large memory consumption, as it contains redundant grid cells for the free space and its size increases exponentially with respect to space dimensionality.



\subsection{Hardware Acceleration of Path Planning} \label{sec:related-hardware-acceleration}
Several works have explored the FPGA and ASIC acceleration of the conventional graph and sampling-based methods, e.g., A*~\cite{YouchangKim16,AtsutakeKosuge20,LukeREverson21,ChengshuoYu21} and RRT~\cite{GurshaantSinghMalik15,SizeXiao16,SizeXiao17,ChiehChung21}.
Kosuge \textit{et al.}~\cite{AtsutakeKosuge20} develops an accelerator for A* graph construction and search on the Xilinx ZCU102.
Since A* operates on grid environments and is subject to the curse of dimensionality, it is challenging to handle higher dimensional cases or larger maps.
For RRT-family algorithms, Malik \textit{et al.}~\cite{GurshaantSinghMalik15} proposes a parallelized architecture for RRT, which first partitions the workspace into grids and distributes them across multiple RRT processes.
This approach involves redundant computations as some RRT processes may explore irrelevant regions for a given problem.
Xiao \textit{et al.}~\cite{SizeXiao16,SizeXiao17} proposes an FPGA accelerator for RRT* that runs tree expansion and rewiring steps in a pipelined manner.
Chung \textit{et al.}~\cite{ChiehChung21} devises a dual-tree RRT with parallel and greedy tree expansion for ASIC implementation.

Some studies leverage GPU~\cite{JoshuaBialkowski11} or distributed computing techniques~\cite{DidierDevaurs13,SamAdeJacobs13} to speed up RRT, while it degrades power efficiency and not suitable for battery-powered edge devices.
In \cite{JoshuaBialkowski11}, collision checks between obstacles and a robot arm are parallelized on GPU; Devaurs \textit{et al.}~\cite{DidierDevaurs13} employs a large-scale distributed memory architecture and a message passing interface.
RRT-family algorithms repeat tree expansion and rewiring steps alternately; they are inherently sequential and difficult to accelerate without a sophisticated technique (e.g., space subdivision, parallel tree expansion).
In contrast, our proposed {\ProposalName} offers more parallelism and is hardware-friendly, as it mainly consists of DNN inferences, and does not operate on complex data structures (e.g., sorting, KNN on K-d trees, graph search).

Only a few works~\cite{KeisukeSugiura22,LingyiHuang22A,LingyiHuang22B} have considered the hardware acceleration of neural planners.
Huang \textit{et al.}~\cite{LingyiHuang22A} presents an accelerator for a sampling-based method with a CNN model, which produces a probability map given an image of the environment for sampling the next robot position.
In \cite{LingyiHuang22B}, an RTL design of a Graph Neural Network-based path explorer rapidly evaluates priority scores for edges in a random geometric graph, and edges with high priority are selected to form a path.
This paper extends our previous work~\cite{KeisukeSugiura22}; instead of only accelerating the DNN inference in MPNet, we implement the whole bidirectional planning algorithm on FPGA.
In addition, we derive a new path planning method, {\ProposalName}, to achieve both higher success rate and speedup.
This paper is one of the first to realize a real-time fully learning-based path planner on a resource-limited FPGA device.

\section{Preliminaries: MPNet} \label{sec:prelim}
In this section, we briefly describe the MPNet~\cite{AhmedQureshi21} algorithm (Alg. \ref{alg:mpnet}), which serves as a basis of our proposal.

\subsection{Notations and Overview of MPNet} \label{sec:prelim-notations-overview}
\begin{figure}
  \centering
  \includegraphics[keepaspectratio, width=0.7\linewidth]{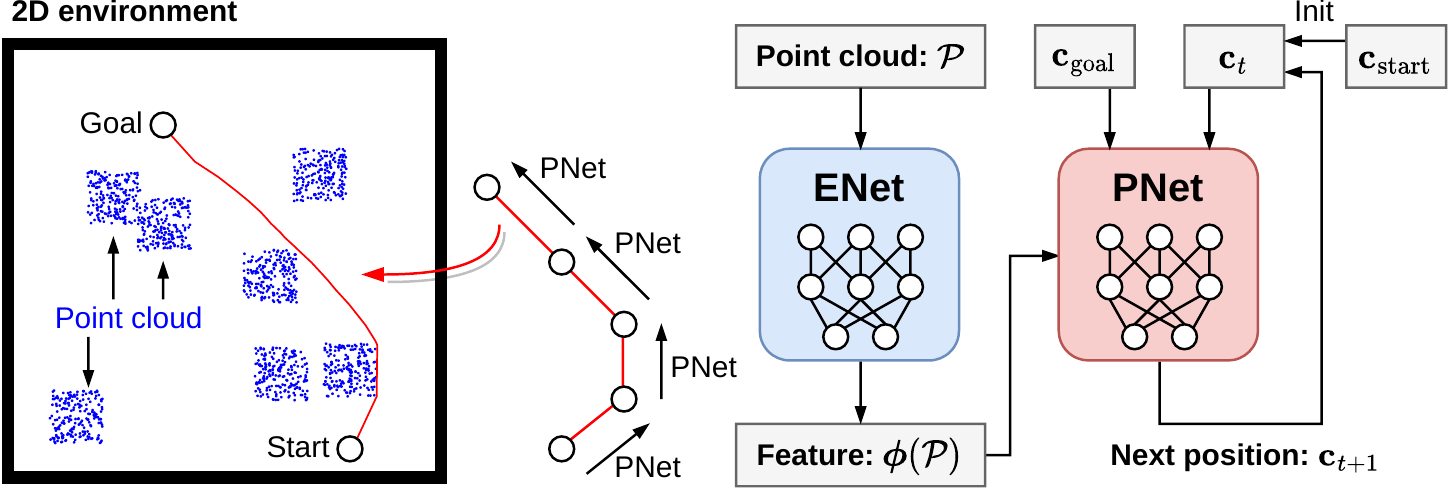}
  \caption{Overview of the MPNet algorithm.
  MPNet takes as input start-goal positions $\mathbf{c}_\mathrm{start}, \mathbf{c}_\mathrm{goal}$ as well as a point cloud $\mathcal{P}$ representing obstacles in the environment (\textcolor{Blue}{blue} points).
  MPNet employs two DNNs, \textbf{ENet} and \textbf{PNet}, for feature extraction and planning.
  ENet extracts a feature $\boldsymbol{\phi}(\mathcal{P})$ from the point cloud, while PNet computes waypoints one at a time and incrementally builds a path from $\mathbf{c}_\mathrm{start}$ to $\mathbf{c}_\mathrm{goal}$.
  Given $\boldsymbol{\phi}(\mathcal{P})$, $\mathbf{c}_\mathrm{goal}$, and a current position $\mathbf{c}_t$ (path endpoint), PNet computes the next position $\mathbf{c}_{t + 1}$ to make a step forward to the goal.}
  \label{fig:mpnet-path-planning}
\end{figure}

Let us consider a robot moving around in a 2D/3D environment $\mathcal{X} \subset \mathbb{R}^D$ ($D = 2, 3$).
For simplicity, the robot is modeled as a point-mass, and its state (configuration) is a position $\mathbf{c} \in \mathbb{R}^D$.
Note that MPNet is a general framework and can be applied to a wide range of problem settings.
Given a pair of start and goal points $\mathbf{c}_\mathrm{start}, \mathbf{c}_\mathrm{goal} \in \mathbb{R}^D$, MPNet tries to find a collision-free path $\tau = \left\{ \mathbf{c}_0, \mathbf{c}_1, \ldots, \mathbf{c}_T \right\}$ ($\mathbf{c}_0, \mathbf{c}_T = \mathbf{c}_\mathrm{start}, \mathbf{c}_\mathrm{goal}$) if exists.
As illustrated in Fig. \ref{fig:mpnet-path-planning} (left), MPNet assumes that obstacle information is represented as a point cloud $\mathcal{P} = \left\{ \mathbf{p}_0, \ldots, \mathbf{p}_{N - 1} \right\} \in \mathbb{R}^{N \times 3}$ containing $N$ points uniformly sampled from the obstacle region.
The notation $\tau \xleftarrow{+} \mathbf{c}$ is a shorthand for $\tau \gets \tau \cup \left\{ \mathbf{c} \right\}$.


Importantly, MPNet uses two DNN models for encoding and planning, namely \textbf{ENet} and \textbf{PNet} (Fig. \ref{fig:mpnet-path-planning} (right)); ENet compresses the obstacle information $\mathcal{P}$ into a feature embedding $\boldsymbol{\phi}(\mathcal{P}) \in \mathbb{R}^M$.
PNet is responsible for sampling the next position $\mathbf{c}_{t + 1}$ which is one step closer to the goal, from the current and goal positions $\mathbf{c}_t, \mathbf{c}_\mathrm{goal}$ as well as the obstacle feature $\boldsymbol{\phi}(\mathcal{P})$.
Fig. \ref{fig:mpnet-path-planning-steps} outlines the algorithm.
MPNet consists of two main steps, referred to as (1) \textbf{initial coarse planning} and (2) \textbf{replanning}, plus (3) a final \textbf{smoothing} step, which are described in the following sections.

\subsection{MPNet Algorithm} \label{sec:prelim-algorithm}
MPNet first extracts a feature $\boldsymbol{\phi}(\mathcal{P})$ (Alg. \ref{alg:mpnet}, line \ref{alg:mpnet-enet}) and proceeds to the \textbf{initial coarse planning} step (line \ref{alg:mpnet-np0}) to roughly plan a path $\tau$ between start and goal points $\mathbf{c}_\mathrm{start}, \mathbf{c}_\mathrm{goal}$.
The bidirectional planning with PNet, referred to as $\mathrm{NeuralPlanner}$ (lines \ref{alg:mpnet-np-begin}-\ref{alg:mpnet-np-end}), plays a central role in this step, which is described as follows.

Given a pair of start-goal points $\mathbf{c}_s, \mathbf{c}_g$, $\mathrm{NeuralPlanner}$ plans two paths $\tau^\mathrm{a}, \tau^\mathrm{b}$ in forward and reverse directions alternately (lines \ref{alg:mpnet-np-alter0}, \ref{alg:mpnet-np-alter1}).
The forward path $\tau^\mathrm{a} = \left\{ \mathbf{c}_s, \ldots, \mathbf{c}_\mathrm{end}^\mathrm{a} \right\}$ is incrementally expanded from start to goal by repeating the PNet inference (lines \ref{alg:mpnet-np-extend0}-\ref{alg:mpnet-np-extend1}).
From the current path endpoint $\mathbf{c}_\mathrm{end}^\mathrm{a}$ and goal $\mathbf{c}_g$, PNet computes a new waypoint $\mathbf{c}_\mathrm{new}^\mathrm{a}$, which becomes a new endpoint of $\tau^\mathrm{a}$ and used as input in the next inference.
Similarly, the backward path $\tau^\mathrm{b} = \left\{ \mathbf{c}_g, \ldots, \mathbf{c}_\mathrm{end}^\mathrm{b} \right\}$ is expanded from goal to start.
In this case, PNet computes a next position $\mathbf{c}_\mathrm{new}^\mathrm{b}$ from an input $\left[ \boldsymbol{\phi}(\mathcal{P}), \mathbf{c}_\mathrm{end}^\mathrm{b}, \mathbf{c}_s \right]$ to get closer to the start position $\mathbf{c}_s$ (lines \ref{alg:mpnet-np-extend2}-\ref{alg:mpnet-np-extend3}).
Note that the newly added edge $(\mathbf{c}_\mathrm{end}^\mathrm{a}, \mathbf{c}_\mathrm{new}^\mathrm{a})$ (or $(\mathbf{c}_\mathrm{end}^\mathrm{b}, \mathbf{c}_\mathrm{new}^\mathrm{b})$) may be in collision; such edge is removed in the later replanning step.
After updating $\tau^\mathrm{a}$ or $\tau^\mathrm{b}$, $\mathrm{NeuralPlanner}$ attempts to connect them and create a path $\tau = \left\{ \mathbf{c}_0, \mathbf{c}_1, \ldots, \mathbf{c}_T \right\}$ between $\mathbf{c}_0, \mathbf{c}_T = \mathbf{c}_s, \mathbf{c}_g$, if there is no obstacle between path endpoints $\mathbf{c}_\mathrm{end}^\mathrm{a}, \mathbf{c}_\mathrm{end}^\mathrm{b}$ (line \ref{alg:mpnet-np-success}).
The above process, i.e., path expansion and collision checking, is repeated until a feasible path is obtained or the maximum number of iterations $I$ is reached.
The algorithm fails if $\tau^\mathrm{a}, \tau^\mathrm{b}$ cannot be connected after $I$ iterations (line \ref{alg:mpnet-np-fail}).

\begin{figure}
  \centering
  \includegraphics[keepaspectratio, width=0.7\linewidth]{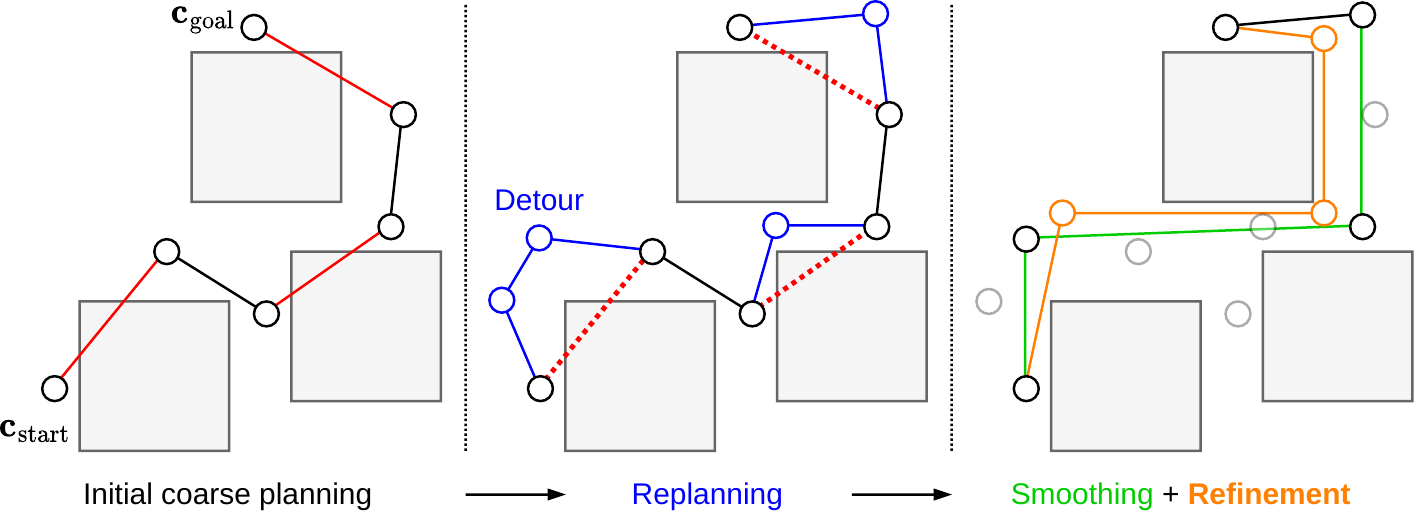}
  \caption{Processing flow of the MPNet path planning.
  MPNet mainly consists of three steps.
  \textbf{initial coarse planning} step (left): MPNet roughly plans a path connecting start-goal points, which may not be collision-free (\textcolor{red}{red} edges).
  \textbf{Replanning} step (center): MPNet removes the edges colliding with obstacles from the initial path and replaces them with alternative edges (detours, \textcolor{blue}{blue}) to obtain a collision-free path.
  \textbf{Smoothing} step (right): The redundant waypoints (semi-transparent points) are pruned to obtain a smooth and straight path (\textcolor{green}{green} edges).
  Our proposed method, \textbf{{\ProposalName}}, performs an additional \textbf{Refinement} step (right) to gradually improve the solution (\textcolor{orange}{orange} edges).}
  \label{fig:mpnet-path-planning-steps}
\end{figure}

The tentative path $\tau$ connecting $\mathbf{c}_\mathrm{start}, \mathbf{c}_\mathrm{goal}$ is obtained from $\mathrm{NeuralPlanner}$ (line \ref{alg:mpnet-np0}); if $\tau$ passes the collision check, then MPNet performs \textbf{smoothing} and returns it as a final solution (lines \ref{alg:mpnet-smooth0}-\ref{alg:mpnet-return0}).
In the smoothing process (Fig. \ref{fig:mpnet-path-planning-steps} (right)), the planner greedily prunes redundant waypoints from $\tau$ to obtain a shorter and smoother path; given three waypoints $\mathbf{c}_i, \mathbf{c}_j, \mathbf{c}_k$ ($i < j < k$), the intermediate one $\mathbf{c}_j$ is dropped if $\mathbf{c}_i$ and $\mathbf{c}_k$ can be directly connected by a straight line.
This involves a collision checking on the new edge $(\mathbf{c}_i, \mathbf{c}_k)$.

As already mentioned, the initial solution $\tau$ may contain edges that collide with obstacles (Fig. \ref{fig:mpnet-path-planning-steps} (left, \textcolor{red}{red} lines)); if this is the case, the planner moves on to the \textbf{replanning} process (lines \ref{alg:mpnet-replan0}, \ref{alg:mpnet-rp-begin}-\ref{alg:mpnet-rp-end}).
For every edge $\mathbf{c}_i, \mathbf{c}_{i + 1} \in \tau$ that is in collision, MPNet tries to plan an alternative sub-path $\tau_{i, i + 1} = \{ \mathbf{c}_i, \mathbf{c}_i^{(1)}, \mathbf{c}_i^{(2)}, \ldots, \mathbf{c}_{i + 1} \}$ between $\mathbf{c}_i, \mathbf{c}_{i + 1}$ to avoid obstacles (Fig. \ref{fig:mpnet-path-planning-steps} (center), line \ref{alg:mpnet-rp-np}).
$\mathrm{NeuralPlanner}$ is again called with $\mathbf{c}_i, \mathbf{c}_{i + 1}$ as input start-goal points.
The replanning fails if $\mathrm{NeuralPlanner}$ cannot plan a detour.
The new intermediate waypoints $\mathbf{c}_i^{(1)}, \mathbf{c}_i^{(2)}, \ldots$ are then inserted to the path (line \ref{alg:mpnet-rp-extend}).
If the resultant path is collision-free, MPNet returns it as a solution after smoothing (lines \ref{alg:mpnet-smooth1}-\ref{alg:mpnet-return1}); otherwise, it runs the replanning again.
In this way, MPNet gradually removes the non-collision-free edges from the solution.
Replanning is repeated for $I_\mathrm{Replan}$ times at maximum, and MPNet fails if a feasible solution is not obtained (line \ref{alg:mpnet-return2}).

One notable feature of MPNet is that, PNet exhibits a stochastic behavior as it utilizes dropout in the inference phase, unlike the typical case where the dropout is only enabled during training.
PNet inference $\left[ \boldsymbol{\phi}(\mathcal{P}), \mathbf{c}_t, \mathbf{c}_\mathrm{goal} \right] \mapsto \mathbf{c}_{t + 1}$ is hence viewed as sampling the next position $\mathbf{c}_{t + 1}$ from a learned distribution $p(\boldsymbol{\phi}(\mathcal{P}), \mathbf{c}_t)$ parameterized by a planning environment $\boldsymbol{\phi}(\mathcal{P})$ and a current position $\mathbf{c}_t$, which represents a promising region around an optimal path from $\mathbf{c}_t$ to $\mathbf{c}_\mathrm{goal}$.
Such dropout sampling (Monte Carlo dropout) was first proposed in \cite{YarinGal16} as an efficient way to estimate the uncertainty in Bayesian Neural Networks (BNNs).
As both $\mathrm{NeuralPlanner}$ and $\mathrm{Replan}$ rely on PNet, they are also non-deterministic and produce different results on the same input.
As a result, MPNet generates different candidate paths in the replanning phase, leading to an increased chance of avoiding obstacles and higher success rate.

\begin{algorithm}[h]
  \small
  \caption{MPNet for 2D and 3D path planning}
  \label{alg:mpnet}
  \begin{algorithmic}[1]
    \Require Start $\mathbf{c}_\mathrm{start}$, goal $\mathbf{c}_\mathrm{goal}$, obstacle point cloud $\mathcal{P}$
    \Ensure Path $\tau = \left\{ \mathbf{c}_0, \mathbf{c}_1, \ldots, \mathbf{c}_T \right\}$
      ($\mathbf{c}_0, \mathbf{c}_T = \mathbf{c}_\mathrm{start}, \mathbf{c}_\mathrm{goal}$)
    \State $\boldsymbol{\phi}(\mathcal{P}) \gets \mathrm{ENet}(\mathcal{P})$
      \Comment{Point cloud feature extraction}
      \vspace{2.5pt} \label{alg:mpnet-enet}

    \LeftComment{\textbf{Initial coarse planning}}
    \State $\tau \gets \mathrm{NeuralPlanner}(\mathbf{c}_\mathrm{start}, \mathbf{c}_\mathrm{goal}, \boldsymbol{\phi}(\mathcal{P}))$
      \label{alg:mpnet-np0}
    \IIf{$\tau = \varnothing$}
      \Return $\varnothing$ \Comment{Failure} \label{alg:mpnet-return3}
    \EndIIf
    \State $\tau \gets \mathrm{Smoothing}(\tau)$ \label{alg:mpnet-smooth0}
    \IIf{$\tau$ is collision-free} \label{alg:mpnet-check0}
      \Return $\tau$ \Comment{Success} \label{alg:mpnet-return0}
    \EndIIf \vspace{2.5pt}

    \LeftComment{\textbf{Replanning}}
    \For{$i = 0, \ldots, I_\mathrm{Replan} - 1$}
      \State $\tau \gets \mathrm{Replan}(\tau, \boldsymbol{\phi}(\mathcal{P}))$ \label{alg:mpnet-replan0}
      \State $\tau \gets \mathrm{Smoothing}(\tau)$ \label{alg:mpnet-smooth1}
      \IIf{$\tau \neq \varnothing$ and $\tau$ is collision-free} \label{alg:mpnet-check1}
        \Return $\tau$ \Comment{Success} \label{alg:mpnet-return1}
      \EndIIf
    \EndFor
    \State \Return $\varnothing$ \Comment{Failure} \vspace{5pt} \label{alg:mpnet-return2}

    \Function{$\mathrm{NeuralPlanner}$}{$\mathbf{c}_s, \mathbf{c}_g, \boldsymbol{\phi}(\mathcal{P})$}
      \label{alg:mpnet-np-begin}
      \State $\tau^\mathrm{a} \gets \left\{ \mathbf{c}_s \right\}, \
        \tau^\mathrm{b} \gets \left\{ \mathbf{c}_g \right\}$
        \Comment{Forward-backward paths}
      \State $\mathbf{c}_\mathrm{end}^\mathrm{a} \gets \mathbf{c}_s, \
        \mathbf{c}_\mathrm{end}^\mathrm{b} \gets \mathbf{c}_g, \ r = 0$
        \Comment{Path endpoints}
      \For{$i = 0, \ldots, I - 1$} \label{alg:mpnet-np-loop}
        \If{$r = 0$} \label{alg:mpnet-np-alter0}
          \Comment{Forward direction (start to goal)}
          \State $\mathbf{c}_\mathrm{new}^\mathrm{a} \gets \mathrm{PNet}(\boldsymbol{\phi}(\mathcal{P}), \mathbf{c}_\mathrm{end}^\mathrm{a}, \mathbf{c}_g)$ \label{alg:mpnet-np-extend0}
          \State $\tau^\mathrm{a} \xleftarrow{+} \left\{ \mathbf{c}_\mathrm{new}^\mathrm{a} \right\}, \
            \mathbf{c}_\mathrm{end}^\mathrm{a} \gets \mathbf{c}_\mathrm{new}^\mathrm{a}, \ r = 1$
            \label{alg:mpnet-np-extend1}
        \ElsIf{$r = 1$} \label{alg:mpnet-np-alter1}
          \Comment{Reverse direction (goal to start)}
          \State $\mathbf{c}_\mathrm{new}^\mathrm{b} \gets \mathrm{PNet}(\boldsymbol{\phi}(\mathcal{P}), \mathbf{c}_\mathrm{end}^\mathrm{b}, \mathbf{c}_s)$ \label{alg:mpnet-np-extend2}
          \State $\tau^\mathrm{b} \xleftarrow{+} \left\{ \mathbf{c}_\mathrm{new}^\mathrm{b} \right\}, \
            \mathbf{c}_\mathrm{end}^\mathrm{b} \gets \mathbf{c}_\mathrm{new}^\mathrm{b}, \ r = 0$
            \label{alg:mpnet-np-extend3}
        \EndIf
        \IIf{$\tau^\mathrm{a}$ and $\tau^\mathrm{b}$ are connectable} \label{alg:mpnet-np-check-connectable}
          \Return $\tau = \left\{ \tau^\mathrm{a}, \tau^\mathrm{b} \right\}$
            \label{alg:mpnet-np-success}
        \EndIIf
      \EndFor
      \State \Return $\varnothing$ \Comment{Failure} \label{alg:mpnet-np-fail}
    \EndFunction \vspace{5pt} \label{alg:mpnet-np-end}

    \Function{$\mathrm{Replan}$}{$\tau = \left\{ \mathbf{c}_0, \ldots, \mathbf{c}_T \right\}, \boldsymbol{\phi}(\mathcal{P})$}
      \label{alg:mpnet-rp-begin}
      \State $\tau_\mathrm{new} \gets \varnothing$
      \For{$i = 0, \ldots, T - 1$}
        \If{$\mathbf{c}_i$ and $\mathbf{c}_{i + 1}$ are connectable} \label{alg:mpnet-rp-check}
          \State $\tau_\mathrm{new} \xleftarrow{+} \left\{ \mathbf{c}_i, \mathbf{c}_{i + 1} \right\}$
        \Else
          \Comment{Plan a detour $\tau_{i, i + 1} = \{ \mathbf{c}_i, \mathbf{c}_i^{(1)}, \mathbf{c}_i^{(2)}, \ldots, \mathbf{c}_{i + 1} \}$}
          \State $\tau_{i, i + 1} \gets \mathrm{NeuralPlanner}(\mathbf{c}_i, \mathbf{c}_{i + 1}, \boldsymbol{\phi}(\mathcal{P}))$ \label{alg:mpnet-rp-np}
          \IIf{$\tau_{i, i + 1} = \varnothing$}
            \Return $\varnothing$ \Comment{Failure} \label{alg:mpnet-rp-fail}
          \EndIIf
          \State $\tau_\mathrm{new} \xleftarrow{+} \tau_{i, i + 1}$ \label{alg:mpnet-rp-extend}
        \EndIf
        \State \Return $\tau_\mathrm{new}$ \Comment{Success}
      \EndFor
    \EndFunction \label{alg:mpnet-rp-end}
  \end{algorithmic}
\end{algorithm}


\section{Method: {\ProposalName}} \label{sec:method}
In this section, we propose a new path planning algorithm, \textbf{{\ProposalName}} (Algs. \ref{alg:p3net}-\ref{alg:p3net-func-batch-plan}), by making improvements to the algorithm and model architecture of MPNet.

\subsection{Algorithmic Improvements} \label{sec:method-algorithmic-improvements}
{\ProposalName} introduces two ideas, (1) \textbf{batch planning} and (2) \textbf{refinement step}, into the MPNet algorithm.
As depicted in Figs. \ref{fig:mpnet-path-planning-steps}--\ref{fig:p3net-batched-planning}, {\ProposalName} (1) estimates multiple paths at the same time to increase the parallelism, and (2) iteratively refines the obtained path to improve its quality.


\subsubsection{Batch Planning} \label{sec:method-batch-planning}
According to the $\mathrm{NeuralPlanner}$ algorithm in MPNet (Alg. \ref{alg:mpnet}, lines \ref{alg:mpnet-np-begin}-\ref{alg:mpnet-np-end}), forward-backward paths $\tau_a, \tau_b$ are incrementally expanded from start-goal points $\mathbf{c}_s, \mathbf{c}_g$ until their endpoints $\mathbf{c}_\mathrm{end}^\mathrm{a}, \mathbf{c}_\mathrm{end}^\mathrm{b}$ are connectable.
In this process, PNet computes a single next position $\mathbf{c}_\mathrm{new}^\mathrm{a}$ ($\mathbf{c}_\mathrm{new}^\mathrm{b}$) from a current endpoint $\mathbf{c}_\mathrm{end}^\mathrm{a}$ ($\mathbf{c}_\mathrm{end}^\mathrm{b}$) and the destination $\mathbf{c}_g$ ($\mathbf{c}_s$) (lines \ref{alg:mpnet-np-extend0}, \ref{alg:mpnet-np-extend2}).
Due to the input batch size of one, PNet cannot fully utilize the parallel computing capability of CPU/GPUs, and also suffers from the kernel launch overheads and frequent data transfers.
To amortize this overhead, two PNet inferences (lines \ref{alg:mpnet-np-extend0}, \ref{alg:mpnet-np-extend2}) can be merged into one with a batch size of two, i.e., two next positions $\left[ \mathbf{c}_s, \mathbf{c}_g \right]$ are computed at once from concatenated inputs $\left[ \mathbf{c}_\mathrm{end}^\mathrm{a}, \mathbf{c}_\mathrm{end}^\mathrm{b} \right], \left[ \mathbf{c}_g, \mathbf{c}_s \right]$.


As shown in Fig. \ref{fig:p3net-batched-planning}, our $\mathrm{NeuralPlannerEx}$ algorithm (Alg. \ref{alg:p3net-func-batch-plan}, lines \ref{alg:p3net-func-np-begin}-\ref{alg:p3net-func-np-end}) takes this idea further by creating a total of $B$ pairs of forward-backward paths (i.e., $\tau_\mathcal{B}^\mathrm{a} = \left[ \tau_0^\mathrm{a}, \ldots, \tau_{B - 1}^\mathrm{a} \right]$, $\tau_\mathcal{B}^\mathrm{b} = \left[ \tau_0^\mathrm{b}, \ldots, \tau_{B - 1}^\mathrm{b} \right]$).
It serves as a drop-in replacement for $\mathrm{NeuralPlanner}$.
It keeps track of the forward-backward path pairs $(\tau_\mathcal{B}^\mathrm{a}, \tau_\mathcal{B}^\mathrm{b})$, which are initialized with start-goal points (lines \ref{alg:p3net-np-init-paths-a}-\ref{alg:p3net-np-init-paths-b}), as well as path lengths $\ell_\mathcal{B}$ (initialized with all ones, line \ref{alg:p3net-np-init-lengths}), path endpoints $\mathbf{C}_\mathcal{B}$, and corresponding destination points $\mathbf{C}_\mathcal{B}^{\mathrm{goal}}$ (lines \ref{alg:p3net-np-init-endpoints}-\ref{alg:p3net-np-init-goals}).


In each iteration $i$, PNet takes $(\boldsymbol{\phi}(\mathcal{P}), \mathbf{C}_\mathcal{B}, \mathbf{C}_\mathcal{B}^{\mathrm{goal}})$ as input and computes the next waypoints $\mathbf{C}_\mathcal{B}^\mathrm{next} = [\mathbf{c}_0^{\mathrm{a}, i + 1}, \mathbf{c}_0^{\mathrm{b}, i + 1}, \ldots, \mathbf{c}_{B - 1}^{\mathrm{a}, i + 1}, \mathbf{c}_{B - 1}^{\mathrm{b}, i + 1}]$ for a batch of paths within one inference step (line \ref{alg:p3net-np-pnet}), resulting in a total batch size of $2B$.
Note that $\mathbf{c}_j^{\mathrm{a}, i + 1}, \mathbf{c}_j^{\mathrm{b}, i + 1}$ denote $i + 1$-th waypoints in $j$-th forward-backward paths ($j \in \left[ 0, B \right)$).
Then, for each sample $j$, the algorithm tries to connect a path pair $\tau_j^\mathrm{a}, \tau_j^\mathrm{b}$, by checking whether any of the three lines connecting $(\mathbf{c}_j^{\mathrm{a}, i + 1}, \mathbf{c}_j^{\mathrm{b}, i})$, $(\mathbf{c}_j^{\mathrm{a}, i}, \mathbf{c}_j^{\mathrm{b}, i + 1})$, and $(\mathbf{c}_j^{\mathrm{a}, i + 1}, \mathbf{c}_j^{\mathrm{b}, i + 1})$ is obstacle-free and hence passable (lines \ref{alg:p3net-np-check-begin}-\ref{alg:p3net-np-check-end}).
If this check passes, $\tau_j^\mathrm{a}$ and $\tau_j^\mathrm{b}$ are concatenated and the result $\tau = \left[ \mathbf{c}_s, \mathbf{c}_s^{\mathrm{a}, 1}, \ldots, \mathbf{c}_s^{\mathrm{b}, 1}, \mathbf{c}_g \right]$ is returned to the caller (line \ref{alg:p3net-np-success}, \textcolor{blue}{blue} path in Fig. \ref{fig:p3net-batched-planning}); otherwise, the algorithm updates the current endpoints $\mathbf{C}_\mathcal{B}$ with the new ones $\mathbf{C}_\mathcal{B}^\mathrm{next}$ and proceeds to the next iteration.
$\mathbf{C}_\mathcal{B}^\mathrm{next}$ is also used to update the paths $\tau_\mathcal{B}^\mathrm{a}, \tau_\mathcal{B}^\mathrm{b}$ accordingly (lines \ref{alg:p3net-np-expand-begin}-\ref{alg:p3net-np-expand-end}).
It fails if no solution is found after the maximum number of iterations $I$.


$\mathrm{NeuralPlannerEx}$ is more likely to find a solution compared to $\mathrm{NeuralPlanner}$, as it creates $B$ candidate paths for a given task.
This allows the replanning process (Alg. \ref{alg:p3net}, lines \ref{alg:p3net-replan-start}-\ref{alg:p3net-replan-end}), which repetitively calls $\mathrm{NeuralPlannerEx}$, to complete in a less number of trials, leading to higher success rates as confirmed in the evaluation (Sec. \ref{sec:eval-success-rate}).
To further improve success rates, {\ProposalName} runs the initial coarse planning for $I_\mathrm{Init} \ge 1$ times (Alg. \ref{alg:p3net}, lines \ref{alg:p3net-np-begin}-\ref{alg:p3net-np-end}), as opposed to MPNet which immediately fails when a feasible path is not obtained in the first attempt (Alg. \ref{alg:mpnet}, lines \ref{alg:mpnet-np0}, \ref{alg:mpnet-return3}).

\subsubsection{Refinement Step} \label{sec:method-refinement-step}
\begin{figure}[htbp]
  \centering
  \includegraphics[keepaspectratio, width=0.5\linewidth]{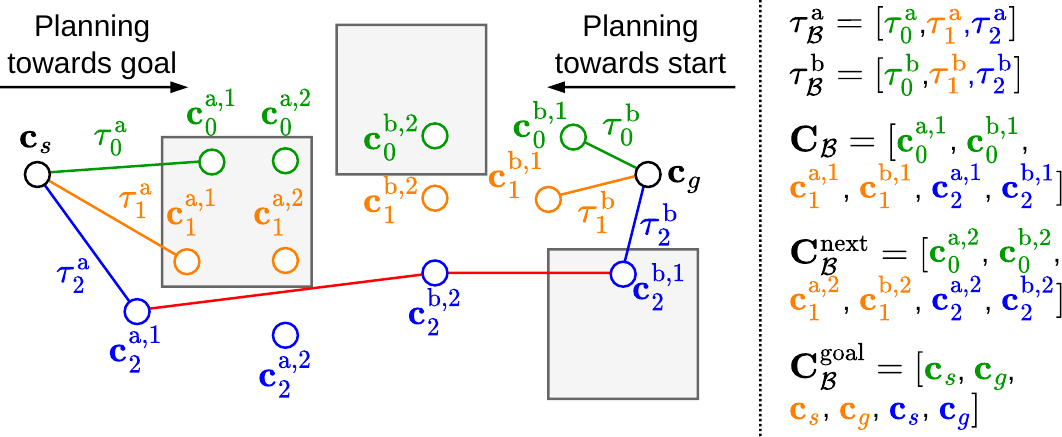}
  \caption{Batch planning in $\mathrm{NeuralPlannerEx}$ algorithm (batch size $B = 3$, iteration $i = 2$).
  The algorithm expands three forward paths $\tau_\mathcal{B}^\mathrm{a} = \left[ \tau_0^\mathrm{a}, \tau_1^\mathrm{a}, \tau_2^\mathrm{a} \right]$ from start $\mathbf{c}_s$ towards goal $\mathbf{c}_g$ and backward paths $\tau_\mathcal{B}^\mathrm{b} = \left[ \tau_0^\mathrm{b}, \tau_1^\mathrm{b}, \tau_2^\mathrm{b} \right]$ from $\mathbf{c}_g$ towards $\mathbf{c}_s$.
  It computes next waypoints $\mathbf{C}_\mathcal{B}^\mathrm{next}$ from the current endpoints $\mathbf{C}_\mathcal{B}$ using PNet, and then tries to connect forward-backward path pairs for each sample.
  In the third sample, the endpoint $\mathbf{c}_2^{\mathrm{a}, 1}$ of forward path and a new waypoint $\mathbf{c}_2^{\mathrm{b}, 2}$ of backward path is connectable; it hence concatenates the forward-backward paths ($\tau_2^\mathrm{a}$, $\tau_2^\mathrm{b}$) to produce a path from start to goal $\tau = \{ \mathbf{c}_s, \mathbf{c}_2^{\mathrm{a}, 1}, \mathbf{c}_2^{\mathrm{b}, 2}, \mathbf{c}_2^{\mathrm{b}, 1}, \mathbf{c}_g \}$.}
  \label{fig:p3net-batched-planning}
\end{figure}

The paths generated by MPNet may not be optimal, since it returns a first found path in an initial coarse planning or a replanning phase.
As highlighted in Alg. \ref{alg:p3net}, lines \ref{alg:p3net-refine-begin}-\ref{alg:p3net-refine-end}, the refinement phase is added at the end of {\ProposalName} to gradually improve the quality of output paths (Fig. \ref{fig:p3net-batched-planning}).
For a fixed number of iterations $I_\mathrm{Refine}$, it computes a new collision-free path $\tau_\mathrm{new}$ based on the current solution $\tau_\mathrm{best}$ (with a cost of $c_\mathrm{best}$) using $\mathrm{Refine}$ algorithm (Alg. \ref{alg:p3net-func-batch-plan}, lines \ref{alg:p3net-refine-func-begin}-\ref{alg:p3net-refine-func-end}), and accepts $\tau_\mathrm{new}$ as a new solution if it lowers the cost ($c_\mathrm{new} < c_\mathrm{best}$).
Same as the replanning phase, $\mathrm{Refine}$ also relies on $\mathrm{NeuralPlannerEx}$ at its core.
It takes the collision-free path $\tau$ as an input and builds a new path $\tau_\mathrm{new}$ as follows: for every edge $(\mathbf{c}_i, \mathbf{c}_{i + 1}) \in \tau$, it plans a path $\tau_{i, i + 1}$ using $\mathrm{NeuralPlannerEx}$ (line \ref{alg:p3net-refine-func-np}) and connects $\mathbf{c}_i, \mathbf{c}_{i + 1}$ with $\tau_{i, i + 1}$ if it is collision-free (line \ref{alg:p3net-refine-func-update}).
The evaluation results (Sec. \ref{sec:eval-path-cost}) confirm that it converges to the optimal solution in most cases.
Note that MPNet can be viewed as a special case of P3Net with $B = 1$, $I_\mathrm{Init} = 1$, and $I_\mathrm{Refine} = 0$.

\begin{algorithm}[h]
  \small
  \caption{{\ProposalName} (changed parts are highlighted in \textcolor{red}{red})}
  \label{alg:p3net}
  \begin{algorithmic}[1]
    \Require Start $\mathbf{c}_\mathrm{start}$, goal $\mathbf{c}_\mathrm{goal}$, obstacle point cloud $\mathcal{P}$
    \Ensure Path $\tau = \left\{ \mathbf{c}_0, \ldots, \mathbf{c}_T \right\}$
      ($\mathbf{c}_0, \mathbf{c}_T = \mathbf{c}_\mathrm{start}, \mathbf{c}_\mathrm{goal}$)
    \State Compute point cloud feature: $\boldsymbol{\phi}(\mathcal{P}) \gets \mathrm{ENet}(\mathcal{P})$
      \vspace{2.5pt} \label{alg:p3net-enet}

    \LeftComment{\textbf{Initial coarse planning}}
    \color{red}
    \For{$i = 0, \ldots, I_\mathrm{Init} - 1$}
      \label{alg:p3net-np-begin}
      \State $\tau \gets \mathrm{NeuralPlannerEx}(\mathbf{c}_\mathrm{start}, \mathbf{c}_\mathrm{goal}, \boldsymbol{\phi}(\mathcal{P}))$
      \IIf{$\tau \neq \varnothing$}
        \AlgBreak \Comment{Success}
      \EndIIf
    \EndFor
    \IIf{$\tau = \varnothing$}
      \Return $\varnothing$ \Comment{Failure}
      \label{alg:p3net-np-end}
    \EndIIf
    \color{black}

    \State $\tau \gets \mathrm{Smoothing}(\tau)$ \label{alg:p3net-smooth0} \vspace{2.5pt}

    \LeftComment{\textbf{Replanning}}
    \If{$\tau$ is not collision-free}
      \label{alg:p3net-replan-start}
      \For{$i = 0, \ldots, I_\mathrm{Replan} - 1$}
        \color{red}
        \State $\tau \gets \mathrm{Replan}(\tau, \boldsymbol{\phi}(\mathcal{P}))$
        \color{black}
        \State $\tau \gets \mathrm{Smoothing}(\tau)$ \label{alg:p3net-smooth1}
        \IIf{$\tau \neq \varnothing$ and $\tau$ is collision-free}
          \AlgBreak \Comment{Success}
        \EndIIf
      \EndFor

      \IIf{$\tau = \varnothing$}
        \Return $\varnothing$ \Comment{Failure}
        \label{alg:p3net-replan-end}
      \EndIIf
    \EndIf \vspace{2.5pt}

    \color{red}
    \LeftComment{\textbf{Refinement}}
    \State $\tau_\mathrm{best} \gets \tau, \ c_\mathrm{best} \gets$ \Call{Cost}{$\tau_\mathrm{best}$}
      \Comment{$\tau$ is now collision-free}
      \label{alg:p3net-cost0} \label{alg:p3net-refine-begin}
    \For{$i = 0, \ldots, I_\mathrm{Refine} - 1$}
      \State $\tau_\mathrm{new} \gets$ \Call{Refine}{$\tau_\mathrm{best}, \boldsymbol{\phi}(\mathcal{P})$}
      \State $\tau_\mathrm{new} \gets$ \Call{Smoothing}{$\tau_\mathrm{new}$} \label{alg:p3net-smooth2}
      \State $c_\mathrm{new} \gets$ \Call{Cost}{$\tau_\mathrm{new}$} \label{alg:p3net-cost1}
      \IIf{$c_\mathrm{new} < c_\mathrm{best}$}
        $c_\mathrm{best} = c_\mathrm{new}, \ \tau_\mathrm{best} = \tau_\mathrm{new}$
          \label{alg:p3net-refine-end}
      \EndIIf
    \EndFor
    \color{black}
    \State \Return $\tau_\mathrm{best}$
  \end{algorithmic}
\end{algorithm}

\begin{algorithm}[h]
  \small
  \caption{Batch Planning and Refinement Step in {\ProposalName}}
  \label{alg:p3net-func-batch-plan}
  \begin{algorithmic}[1]
    \Function{$\mathrm{NeuralPlannerEx}$}{$\mathbf{c}_s, \mathbf{c}_g, \boldsymbol{\phi}(\mathcal{P})$}
      \label{alg:p3net-func-np-begin}
      \LeftComment{Initialize batch of current and goal positions}
      \State $\mathbf{C}_\mathcal{B} \gets \left[
        \mathbf{c}_0^\mathrm{a, 0}, \mathbf{c}_0^\mathrm{b, 0}, \ldots,
        \mathbf{c}_{B - 1}^\mathrm{a, 0}, \mathbf{c}_{B - 1}^\mathrm{b, 0} \right] \in \mathbb{R}^{2B \times D}$
        \label{alg:p3net-np-init-endpoints}
      \Statex \hspace*{\algorithmicindent} \hspace{5pt}
        ($\forall j \ \mathbf{c}_j^\mathrm{a, 0} = \mathbf{c}_s, \mathbf{c}_j^\mathrm{b, 0} = \mathbf{c}_g$)
      \State $\mathbf{C}_\mathcal{B}^\mathrm{goal} \gets \left[
        \mathbf{c}_g, \mathbf{c}_s, \mathbf{c}_g, \mathbf{c}_s, \ldots,
        \mathbf{c}_g, \mathbf{c}_s \right] \in \mathbb{R}^{2B \times D}$ \vspace{2.5pt}
        \label{alg:p3net-np-init-goals}

      \LeftComment{Initialize batch of paths and lengths}
      \State $\tau_\mathcal{B}^\mathrm{a} \gets \left[
        \tau_0^\mathrm{a}, \tau_1^\mathrm{a}, \ldots, \tau_{B - 1}^\mathrm{a} \right], \
        \forall j \ \tau_j^\mathrm{a} = \left\{ \mathbf{c}_s \right\}$
        \label{alg:p3net-np-init-paths-a}
      \State $\tau_\mathcal{B}^\mathrm{b} \gets \left[
        \tau_0^\mathrm{b}, \tau_1^\mathrm{b}, \ldots, \tau_{B - 1}^\mathrm{b} \right], \
        \forall j \ \tau_j^\mathrm{b} = \left\{ \mathbf{c}_g \right\}$
        \label{alg:p3net-np-init-paths-b}
      \State $\ell_\mathcal{B} \gets \left[ \ell_0^\mathrm{a}, \ell_0^\mathrm{b}, \ldots,
        \ell_{B - 1}^\mathrm{a}, \ell_{B - 1}^\mathrm{b} \right], \
        \forall j \ \ell_j^\mathrm{x} = |\tau_j^\mathrm{x}| = 1$ \vspace{2.5pt}
        \label{alg:p3net-np-init-lengths}

      \For{$i = 0, \ldots, I - 1$}
        \State $\mathbf{C}_\mathcal{B}^\mathrm{next} \gets$ \Call{PNet}{$
          \boldsymbol{\phi}(\mathcal{P}), \mathbf{C}_\mathcal{B},
          \mathbf{C}_\mathcal{B}^\mathrm{goal}$} \Comment{Next positions}
        \label{alg:p3net-np-pnet}
        \Statex \hspace*{\algorithmicindent} \hspace{5pt}
          ($\mathbf{C}_\mathcal{B}^\mathrm{next} = \left[
            \mathbf{c}_0^{\mathrm{a}, i + 1}, \mathbf{c}_0^{\mathrm{b}, i + 1}, \ldots,
            \mathbf{c}_{B - 1}^{\mathrm{a}, i + 1}, \mathbf{c}_{B - 1}^{\mathrm{b}, i + 1} \right]$)
        \For{$j = 0, \ldots, B - 1$} \Comment{Collision checks}
          \label{alg:p3net-np-check-begin}
          \If{($\mathbf{c}_j^{\mathrm{a}, i + 1}, \mathbf{c}_j^{\mathrm{b}, i}$) are connectable}
            \State $\mathrm{expandA} \gets 1, \ \mathrm{expandB} \gets 0, \ s \gets 1$
          \ElsIf{($\mathbf{c}_j^{\mathrm{a}, i}, \mathbf{c}_j^{\mathrm{b}, i + 1}$) are connectable}
            \State $\mathrm{expandA} \gets 0, \ \mathrm{expandB} \gets 1, \ s \gets 1$
          \ElsIf{($\mathbf{c}_j^{\mathrm{a}, i + 1}, \mathbf{c}_j^{\mathrm{b}, i + 1}$) are connectable}
            \State $\mathrm{expandA} \gets 1, \ \mathrm{expandB} \gets 1, \ s \gets 1$
          \Else
            \State $\mathrm{expandA} \gets 1, \ \mathrm{expandB} \gets 1, \ s \gets 0$
            \label{alg:p3net-np-check-end}
          \EndIf

          \IIf{$\mathrm{expandA}$} \label{alg:p3net-np-expand-begin}
            $\tau_j^\mathrm{a} \xleftarrow{+} \{ \mathbf{c}_j^{\mathrm{a}, i + 1} \}, \
            \ell_j^\mathrm{a} \gets \ell_j^\mathrm{a} + 1$
          \EndIIf

          \IIf{$\mathrm{expandB}$}
            $\tau_j^\mathrm{b} \xleftarrow{+} \{ \mathbf{c}_j^{\mathrm{b}, i + 1} \}, \
            \ell_j^\mathrm{b} \gets \ell_j^\mathrm{b} + 1$
          \EndIIf \label{alg:p3net-np-expand-end}

          \IIf{$s = 1$}
            \Return $\tau = \left\{ \tau_j^\mathrm{a}, \tau_j^\mathrm{b} \right\}$
              \Comment{Success} \label{alg:p3net-np-success}
          \EndIIf

          \State $\mathbf{C}_\mathcal{B} \gets \mathbf{C}_\mathcal{B}^\mathrm{next}$
        \EndFor
      \EndFor
      \State \Return $\varnothing$ \Comment{Failure}
    \EndFunction \vspace{5pt} \label{alg:p3net-func-np-end}


    \Function{$\mathrm{Refine}$}{$\tau = \left\{ \mathbf{c}_0, \ldots, \mathbf{c}_T \right\}, \boldsymbol{\phi}(\mathcal{P})$}
      \label{alg:p3net-refine-func-begin}
      \State $\tau_\mathrm{new} \gets \varnothing$
      \For{$i = 0, \ldots, T - 1$}
        \State $\tau_{i, i + 1} \gets$ \Call{NeuralPlannerEx}{$\mathbf{c}_i, \mathbf{c}_{i + 1}, \boldsymbol{\phi}(\mathcal{P})$}
        \label{alg:p3net-refine-func-np} \\
        \Comment{Compute a new path connecting $\mathbf{c}_i$ and $\mathbf{c}_{i + 1}$}
        \If{$\tau_{i, i + 1} \neq \varnothing$ and $\tau_{i, i + 1}$ is collision-free}
          \label{alg:p3net-refine-func-check}
          \State $\tau_\mathrm{new} \xleftarrow{+} \tau_{i, i + 1}$
          \Comment{Use new path} \label{alg:p3net-refine-func-update}
        \Else
          \State $\tau_\mathrm{new} \xleftarrow{+} \left\{ \mathbf{c}_i, \mathbf{c}_{i + 1} \right\}$
            \Comment{Use current solution}
        \EndIf
      \EndFor
      \State \Return $\tau_\mathrm{new}$
    \EndFunction \label{alg:p3net-refine-func-end}
  \end{algorithmic}
\end{algorithm}

\subsection{Lightweight Encoding and Planning Networks} \label{sec:method-improved-models}
As depicted in Fig. \ref{fig:mpnet-path-planning} (right), encoding and planning networks ({\EPNet}) are employed in the MPNet framework to extract features from obstacle point clouds and progressively compute waypoints on the output paths.
Instead of {\EPNet}, {\ProposalName} uses a lightweight encoder with a PointNet~\cite{CharlesQi17} backbone (\textbf{{\NewENet}}) to extract more robust features, in conjunction with a downsized planning network (\textbf{{\NewPNet}}) for better parameter efficiency and faster inference, which are described in the following subsections\footnote{To distinguish the models for 2D/3D planning tasks, we suffix them with -\textbf{2D}/-\textbf{3D} when necessary. A fully-connected (FC) layer with $m$ input and $n$ output channels is denoted as $\mathrm{FC}(m, n)$, a 1D batch normalization with $n$ channels as $\mathrm{BN}(n)$, a 1D max-pooling with window size $n$ as $\mathrm{MaxPool}(n)$, and a dropout with a rate $p \in \left[ 0, 1 \right)$ as $\mathrm{Dropout}(p)$, respectively.}.

\begin{figure}[h]
  \centering
  \includegraphics[keepaspectratio, width=\linewidth]{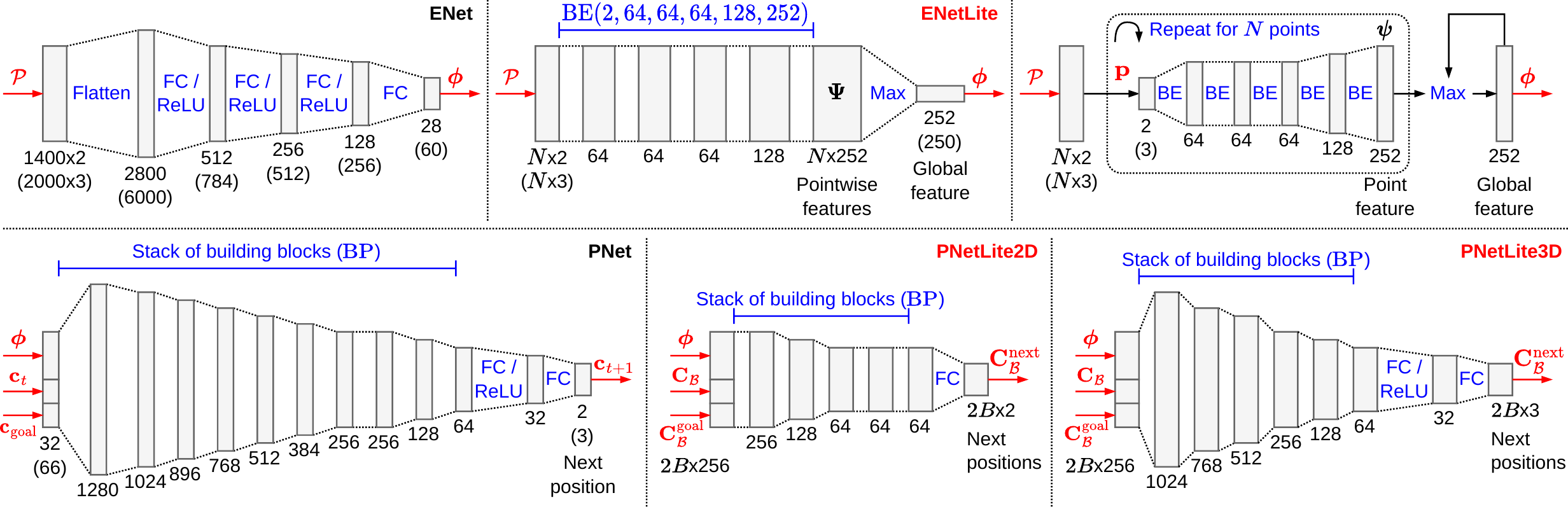}
  \caption{Encoding and planning networks for MPNet and {\ProposalName} (top right: sequential feature extraction in Sec. \ref{sec:impl-encoder-update}).
  $\mathrm{BE}$ is a shorthand for a building block consisting of a fully-connected layer, batch normalization, and ReLU activation.
  $\mathrm{BP}$ is a shorthand for a building block consisting of a fully-connected layer, ReLU activation, and a dropout with a probability of $0.5$.}
  \label{fig:enet-and-pnet-model}
\end{figure}



\subsubsection{{\NewENet}: PointNet-based Encoding Network} \label{sec:method-improved-enet}
As shown in Fig. \ref{fig:enet-and-pnet-model} (top left), MPNet uses a simple encoder architecture (ENet) stacking four FC layers.
It directly processes raw point coordinates, and hence costly preprocessing such as normal estimation or clustering is not required.
ENet2D takes a point cloud containing 1,400 representative points residing on the obstacles $\mathcal{P} \in \mathbb{R}^{1,400 \times 2}$, flattens it into a 2,800D vector, and produces a 28D feature vector $\boldsymbol{\phi}(\mathcal{P}) \in \mathbb{R}^{28}$ in a single forward pass.
Similarly, ENet3D extracts a 60D feature vector $\boldsymbol{\phi}(\mathcal{P})$ from a 3D point cloud of size 2,000 with a series of FC layers\footnote{ENet3D is denoted as $\mathrm{FC}(6000, 784) \to \mathrm{ReLU} \to \mathrm{FC}(512) \to \mathrm{ReLU} \to \mathrm{FC}(256) \to \mathrm{ReLU} \to \mathrm{FC}(60)$.}.


In spite of its simplicity, ENet has the following major drawbacks; (1) the number of input points is fixed to 1,400 or 2,000 regardless of the complexity of planning environments, (2) the number of parameters grows linearly with the number of points, and more importantly, (3) the output feature is affected by the input orderings.
This means that ENet produces a different feature if any of the two points are swapped; since the input still represents exactly the same point set, the result should remain unchanged.


{\ProposalName} avoids these drawbacks by using PointNet~\cite{CharlesQi17} as an encoder backbone, referred to as \textbf{{\NewENet}}.
PointNet is specifically designed for point cloud processing, and Fig. \ref{fig:enet-and-pnet-model} (top center) presents its architecture.
It is still a fully-connected network and directly operates on raw point clouds.
{\NewENet}2D first extracts 252D individual features $\left\{ \boldsymbol{\psi}(\mathbf{p}_0), \ldots, \boldsymbol{\psi}(\mathbf{p}_{N - 1}) \right\}$ for each point using a set of blocks represented as $\mathrm{BE}(2, 64, 64, 64, 128, 252)$\footnote{$\mathrm{BE}(m, n) = \mathrm{FC}(m, n) \to \mathrm{BN}(n) \to \mathrm{ReLU}$ is a basic building block that maps $m$D point features into an $n$D space.}.
It then computes a 252D global feature $\boldsymbol{\phi}(\mathcal{P}) = \max\left(\boldsymbol{\psi}(\mathbf{p}_0), \ldots, \boldsymbol{\psi}(\mathbf{p}_{N - 1})\right)$ by aggregating these pointwise features via max-pooling.
{\NewENet}3D has the same structure as in the 2D case, except the first and the last building blocks are replaced with $\mathrm{BE}(3, 64)$ and $\mathrm{BE}(250)$, respectively, to extract 250D features from 3D point clouds.


Compared to ENet, {\NewENet} can handle point clouds of any size, and the number of parameters is independent from the input size.
{\NewENet}2D(3D) provides informative features with 9x (28/252) and 4.17x (60/250) more dimensions, while requiring 31.73x (1.60M/0.05M) and 104.47x (5.25M/0.05M) less parameters than ENet2D(3D).
PointNet processes each point in a point cloud sequentially and thus avoids random accesses.
In addition, as {\NewENet} involves only pointwise operations and a symmetric pooling function, its output $\boldsymbol{\phi}(\mathcal{P})$ is invariant to the permutation of input points, leading to better training efficiency and robustness.

\subsubsection{{\NewPNet}: Lightweight Planning Network} \label{sec:method-improved-pnet}
The original PNet is formed by a set of building blocks\footnote{$\mathrm{BP}(m, n) = \mathrm{FC}(m, n) \to \mathrm{ReLU} \to \mathrm{Dropout}(0.5)$.}, as shown in Fig. \ref{fig:enet-and-pnet-model} (bottom left).
It takes a concatenated input $\left[ \boldsymbol{\phi}(\mathcal{P}), \mathbf{c}_t, \mathbf{c}_\mathrm{goal} \right]$ consisting of an obstacle feature $\boldsymbol{\phi}(\mathcal{P})$ passed from ENet, a current position $\mathbf{c}_t$, and a destination $\mathbf{c}_\mathrm{goal}$, and computes the next position $\mathbf{c}_{t + 1}$ which is one step closer to $\mathbf{c}_\mathrm{goal}$.
Notably, PNet2D/3D have the same set of hidden layers; the only difference is in the leading and trailing layers.
PNet2D uses $\mathrm{BP}(32, 1280)$ and $\mathrm{FC}(2)$ to produce 2D coordinates from $28 + 2 \cdot 2 = 32$D inputs, whereas PNet3D uses $\mathrm{BP}(66, 1280)$ and $\mathrm{FC}(3)$ to handle $60 + 3 \cdot 2 = 66$D inputs and $3$D outputs.


Such design has a problem of low parameter efficiency especially in the 2D case; PNet2D will contain redundant layers which do not contribute to the successful planning and only increase the inference time.
The network architecture can be adjusted to the number of state dimensions and the complexity of planning tasks.
In addition, as discussed in Sec. \ref{sec:method-improved-enet}, PointNet encoder provides robust (permutation-invariant) features which better represent the planning environment.
Assuming that MPNet uses a larger PNet in order to compensate for the lack of robustness and geometric information in ENet-extracted features, it is reasonable to expect that PointNet allows the use of more shallower networks for path planning.


From the above considerations, {\ProposalName} employs more compact planning networks with fewer building blocks, \textbf{{\NewPNet}}.
{\NewPNet}2D (Fig. \ref{fig:enet-and-pnet-model} (bottom center)) is composed of six building blocks\footnote{{\NewPNet}2D is denoted as $\mathrm{BP}(256, 256, 128, 64, 64, 64, 2)$.} to compute a 2D position from a $252 + 2 \cdot 2 = 256$D input.
As described in Sec. \ref{sec:method-batch-planning}, $\mathrm{NeuralPlannerEx}$ plans $B$ pairs of forward-backward paths and thus the input batch size increases to $2B$; {\NewPNet} computes a batch of next positions $\mathbf{C}_\mathcal{B}^\mathrm{next} \in \mathbb{R}^{2B \times 2}$ from a matrix $[\boldsymbol{\phi}(\mathcal{P}), \mathbf{C}_\mathcal{B}^\mathrm{next}, \mathbf{C}_\mathcal{B}^\mathrm{goal}] \in \mathbb{R}^{2B \times 256}$.
{\NewPNet}3D is obtained by removing a few blocks from PNet3D and replacing the first block with $\mathrm{BP}(256, 1024)$ to process $250 + 3 \cdot 2 = 256$D inputs, as depicted in Fig. \ref{fig:enet-and-pnet-model} (bottom right).
{\NewPNet}2D(3D) has 32.58x (3.76M/0.12M) and 2.35x (3.80M/1.62M) less parameters than PNet2D(3D); combining the results from Sec. \ref{sec:method-improved-enet}, {\NewEPNet} together achieves 32.32x (5.43x) parameter reduction in the 2D (3D) case.

\section{Implementation} \label{sec:impl}
This section details the design and implementation of \textbf{{\IPName}}, a custom IP core for {\ProposalName}.
It has three submodules, namely, (1) \textbf{Encoder}, (2) \textbf{NeuralPlanner}, and (3) \textbf{CollisionChecker} (Fig. \ref{fig:board-level-impl}), which cover most of the {\ProposalName} (Alg. \ref{alg:p3net}) except the evaluation of path costs (lines \ref{alg:p3net-cost0}, \ref{alg:p3net-cost1}).

\subsection{Encoder Module} \label{sec:impl-encoder}
While the number of parameter is greatly reduced, {\NewENet} requires a longer inference time than ENet (Table \ref{tbl:inference-collision-check-time} (top)), as it extracts local features for each point, which amounts to $N$ times of forward pass.
\textbf{Encoder} module is to accelerate the {\NewENet} inference (Alg. \ref{alg:p3net}, line \ref{alg:p3net-enet}).
To reduce the memory cost from $\mathcal{O}(N)$ to $\mathcal{O}(1)$, it sequentially updates the output feature $\boldsymbol{\phi}(\mathcal{P})$.
In addition, it applies both coarse and fine-grained optimizations.

\subsubsection{Memory-Efficient Sequential Feature Extraction} \label{sec:impl-encoder-update}
The typical approach for extracting an {\NewENet} feature is to first compute individual features $\mathbf{\Psi} = \left\{ \boldsymbol{\psi}(\mathbf{p}_0), \ldots, \boldsymbol{\psi}(\mathbf{p}_{N - 1}) \right\}$ for all points in one shot, creating a matrix of size $(N, 252)$ (or $(N, 250)$ in the 3D case), and apply max-pooling over $N$ features to produce a 252D (250D) global feature $\boldsymbol{\phi}(\mathcal{P}) = \max(\mathbf{\Psi})$.
In this case, each building block involves a matrix-matrix operation\footnote{Each building block is denoted as $\mathrm{BE}(m, n) = \mathrm{ReLU}(\mathrm{BN}(\mathbf{X} \mathbf{W} + \mathbf{1} \mathbf{b}^\top))$, where $\mathbf{X} \in \mathbb{R}^{N \times m}$ is a stack of $m$D features, $\mathbf{W} \in \mathbb{R}^{m \times n}, \mathbf{b} \in \mathbb{R}^n$ are weight and bias parameters of a FC layer, and $\mathbf{1} \in \mathbb{R}^N$ is a vector of ones, respectively.}.
While it offers a high degree of parallelism and are hardware-amenable, the buffer size for layer outputs is $\mathcal{O}(N)$, which incurs a high utilization of the scarce on-chip memory.

Since the operations for each point is independent except the last max-pooling, instead of following the above approach, the module computes a point feature $\boldsymbol{\psi}(\mathbf{p}_i)$ one-by-one and updates the global feature by taking a maximum $\boldsymbol{\phi}(\mathcal{P}) \gets \max(\boldsymbol{\phi}(\mathcal{P}), \boldsymbol{\psi}(\mathbf{p}_i))$ (Fig. \ref{fig:enet-and-pnet-model} (top right)).
After repeating this process for all $N$ points, the result $\boldsymbol{\phi}(\mathcal{P})$ is exactly the same as the previous one, $\max(\mathbf{\Psi})$.
In this way, the computation inside each building block turns into a matrix-vector operation\footnote{$\mathrm{BE}(m, n) = \mathrm{ReLU}(\mathrm{BN}(\mathbf{W}^\top \mathbf{x} + \mathbf{b}))$.}.
As it only requires an input and output buffer for a single point, the buffer size is reduced from $\mathcal{O}(N)$ to $\mathcal{O}(1)$; Encoder module can now handle point clouds of any size regardless of the limited on-chip memory.


\subsubsection{{\NewENet} Inference} \label{sec:impl-encoder-inference}
Encoder module consists of three kinds of submodules: $\mathrm{FC}(m, n)$, $\mathrm{BN\text{-}ReLU}(n)$, and $\mathrm{Max}$.
$\mathrm{FC}(m, n)$ involves a matrix-vector product $\mathbf{W}^\top \mathbf{x} + \mathbf{b}$, and $\mathrm{Max}$ updates the feature by $\max(\boldsymbol{\phi}(\mathcal{P}), \boldsymbol{\psi})$.
$\mathrm{BN\text{-}ReLU}(n)$ couples the batch normalization and ReLU.
It is written as $\max(\mathbf{0}, (\mathbf{x} - \boldsymbol{\mu}_\mathcal{B}) \cdot \mathbf{s} + \boldsymbol{\beta}), \mathbf{s} = \boldsymbol{\gamma} / \sqrt{\boldsymbol{\sigma}_\mathcal{B}^2 + \varepsilon} \in \mathbb{R}^n$ with a little abuse of notations\footnote{$\max(\mathbf{0}, \cdot)$ corresponds to ReLU, $\boldsymbol{\mu}_\mathcal{B}, \boldsymbol{\sigma}_\mathcal{B}^2 \in \mathbb{R}^n$ denote the mean and standard deviation estimated from training data, $\boldsymbol{\gamma}, \boldsymbol{\beta} \in \mathbb{R}^n$ are the learned weight and bias, and $\varepsilon > 0$ is a small positive value to prevent zero division, respectively. The scale, $\mathbf{s}$, is precomputed on the CPU instead of keeping individual parameters ($\boldsymbol{\gamma}$, $\boldsymbol{\sigma}_\mathcal{B}^2$), which removes some operations (e.g., square root and division).}.
$\mathrm{FC}(m, n)$ and $\mathrm{BN\text{-}ReLU}(n)$ are parallelized by partially unrolling the loop and partitioning the relevant buffers.
Besides, a dataflow optimization is applied for the fully-pipelined execution of these submodules (Fig. \ref{fig:module-encoder} (top)).
This effectively reduces the latency while requiring a few additional memory resources for inserting pipeline buffers.




\subsubsection{Processing Flow of Encoder Module} \label{sec:impl-encoder-flow}
\begin{figure}[htbp]
  \centering
  \includegraphics[keepaspectratio, width=0.55\linewidth]{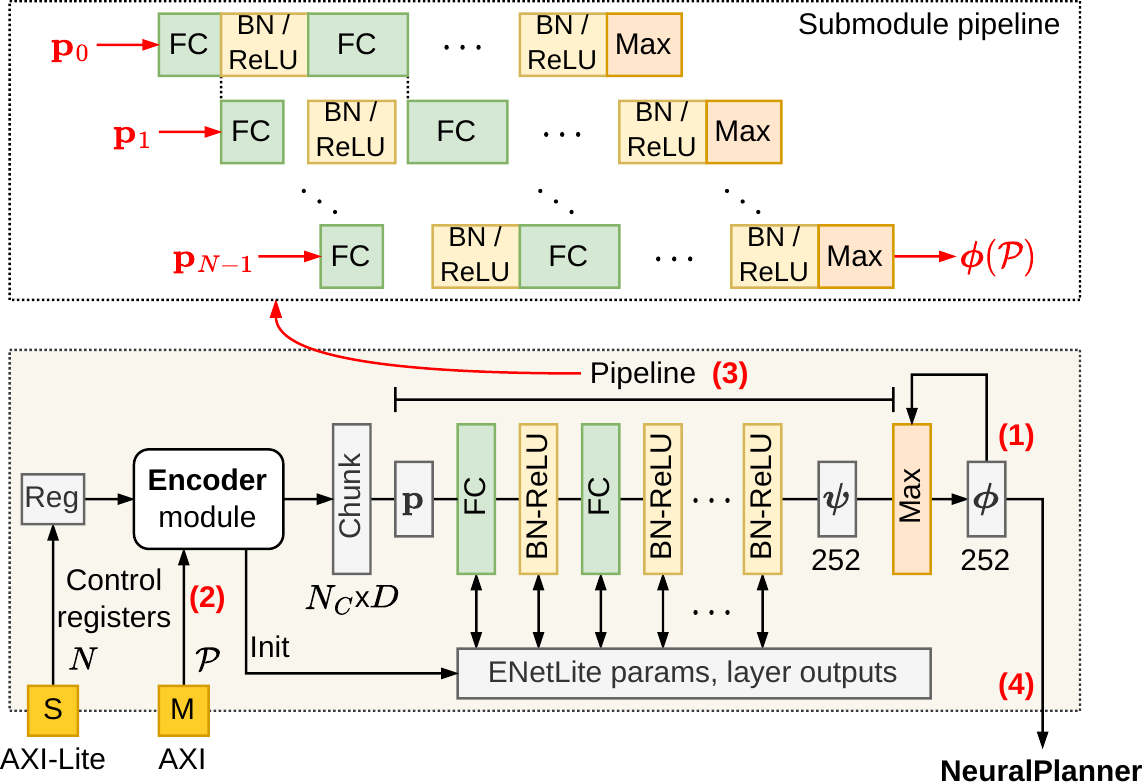}
  \caption{Pipeline of submodules for {\NewENet} (top) and block diagram of Encoder module (bottom).}
  \label{fig:module-encoder}
\end{figure}

Fig. \ref{fig:module-encoder} (bottom) shows the block diagram of Encoder module.
Upon a request from the CPU, it first (1) initializes the output feature $\boldsymbol{\phi}(\mathcal{P})$ with zeros.
Then, the input $\mathcal{P}$ is processed in fixed-size chunks as follows: the module (2) transfers a chunk from DRAM to a point buffer of size $(N_C, D)$ ($N_C = 64, D = 2, 3$) via burst transfer, and (3) the submodule pipeline consumes points in this buffer one-by-one to update the output.
After repeating this process for all $\lceil N / N_C \rceil$ chunks, (4) the output $\boldsymbol{\phi}(\mathcal{P})$ is written to the on-chip buffer of size $(2B, 256)$ for later use in NeuralPlanner module.
On-chip buffers for an input chunk, an output $\boldsymbol{\phi}$, intermediate outputs, and model parameters are all implemented with BRAM.
The lightweight {\NewENet} model and a sequential update of outputs, which substantially reduce the model parameters and make buffer sizes independent from the number of input points, are essential to place the buffers on BRAM and avoid DRAM accesses during computation.


\subsection{NeuralPlanner Module} \label{sec:impl-neural-planner}
As apparent in Algs. \ref{alg:mpnet}--\ref{alg:p3net}, the bidirectional neural planning ($\mathrm{NeuralPlannerEx}$) is at the core of {\ProposalName} and thus has a significant impact on the overall performance.
In contrast to the previous work~\cite{KeisukeSugiura22}, which only offloads the PNet inference (line \ref{alg:p3net-np-pnet}) to the custom IP core, \textbf{NeuralPlanner} module runs the entire $\mathrm{NeuralPlannerEx}$ algorithm (Alg. \ref{alg:p3net-func-batch-plan}).
Considering that it alternates between {\NewPNet} inference and collision checks (lines \ref{alg:p3net-np-check-begin}-\ref{alg:p3net-np-check-end}), implementing both in one place eliminates the unnecessary CPU-FPGA communication and greatly improves the speed.

\subsubsection{{\NewPNet} Inference} \label{sec:impl-neural-planner-inference}
For {\NewPNet} inference, the module contains two types of submodules: $\mathrm{FC}(m, n)$ and $\mathrm{Dropout\text{-}ReLU}(p) = \mathrm{ReLU} \to \mathrm{Dropout}(p)$ which fuses ReLU and dropout into a single pipelined loop.
$\mathrm{FC}(m, n)$ exploits the fine-grained (data-level) parallelism as described in Sec. \ref{sec:impl-encoder-inference}.
Dropout is the key for the stochastic behavior, and PNet inference is interpreted as a deeply-informed sampling (Sec. \ref{sec:prelim-algorithm}).
$\mathrm{Dropout\text{-}ReLU}(p)$ with $p = 0.5$ is implemented using a famous Mersenne-Twister (MT)~\cite{MakotoMatsumoto98}; it first generates 32-bit pseudorandom numbers $\left[ r_{bi} \right] \in \mathbb{N}^{2B \times n}$ for a batch input $\mathbf{X} = \left[ x_{bi} \right] \in \mathbb{R}^{2B \times n}$, and replaces $x_{bi}$ with zero if $x_{bi} < 0$ (ReLU) or $r_{bi} < 2^{31}$ (the maximum 32-bit integer multiplied by $p$).
Note that in the 3D case, parameters for the first three FC layers are kept on the DRAM due to the limited on-chip memory resources, which necessitates a single sweep (burst transfer) of weight and bias parameters in a forward pass.



\subsubsection{Collision Checking} \label{sec:impl-neural-planner-collision-check}
In addition to {\NewPNet} inference, NeuralPlanner module deals with collision checking.
It adopts a simple approach based on the discretization~\cite{JoshuaBialkowski11,AhmedQureshi21} to check whether the lines are in collision.
The line between $\mathbf{c}^\mathrm{start}, \mathbf{c}^\mathrm{end}$ is divided into segments with a predefined interval $\delta$ (Fig. \ref{fig:module-planner} (right top)), producing equally spaced midpoints $\mathbf{c}_0, \ldots, \mathbf{c}_M$\footnote{$\mathbf{c}_i = \mathbf{c}^\mathrm{start} + (i / M) \mathbf{\Delta}$, $\mathbf{\Delta} = \mathbf{c}^\mathrm{end} - \mathbf{c}^\mathrm{start}$, $M = \left\| \mathbf{\Delta} \right\| / \delta$.}, and then each midpoint $\mathbf{c}_i$ is tested.
If any midpoint collides with any obstacle, the line is in collision.
To simplify the implementation, the module assumes that each obstacle is rectangular and represented as a bounding box with minimum and maximum corner points $\mathbf{c}_{i, \min}^\mathrm{obs}, \mathbf{c}_{i, \max}^\mathrm{obs}$.
The module contains eight $\mathrm{Check}$ submodules to test a midpoint $\mathbf{c}_i$ with eight obstacles in parallel (Fig. \ref{fig:module-planner} (right bottom)).


\subsubsection{Processing Flow of NeuralPlanner Module} \label{sec:impl-neural-planner-flow}
\begin{figure}[htbp]
  \centering
  \includegraphics[keepaspectratio, width=0.7\linewidth]{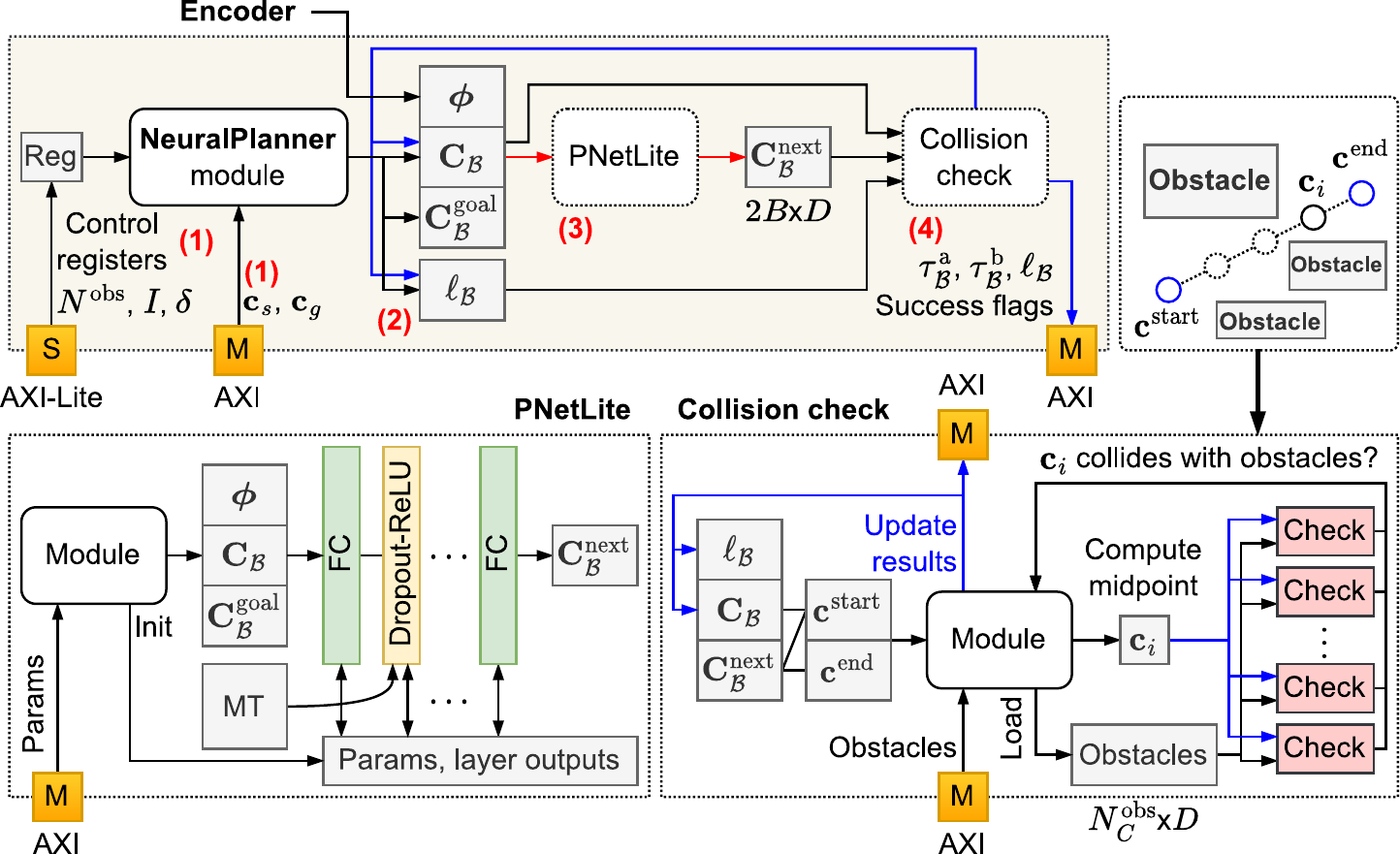}
  \caption{Block diagram of NeuralPlanner module.}
  \label{fig:module-planner}
\end{figure}

\begin{figure}[htbp]
  \centering
  \includegraphics[keepaspectratio, width=0.6\linewidth]{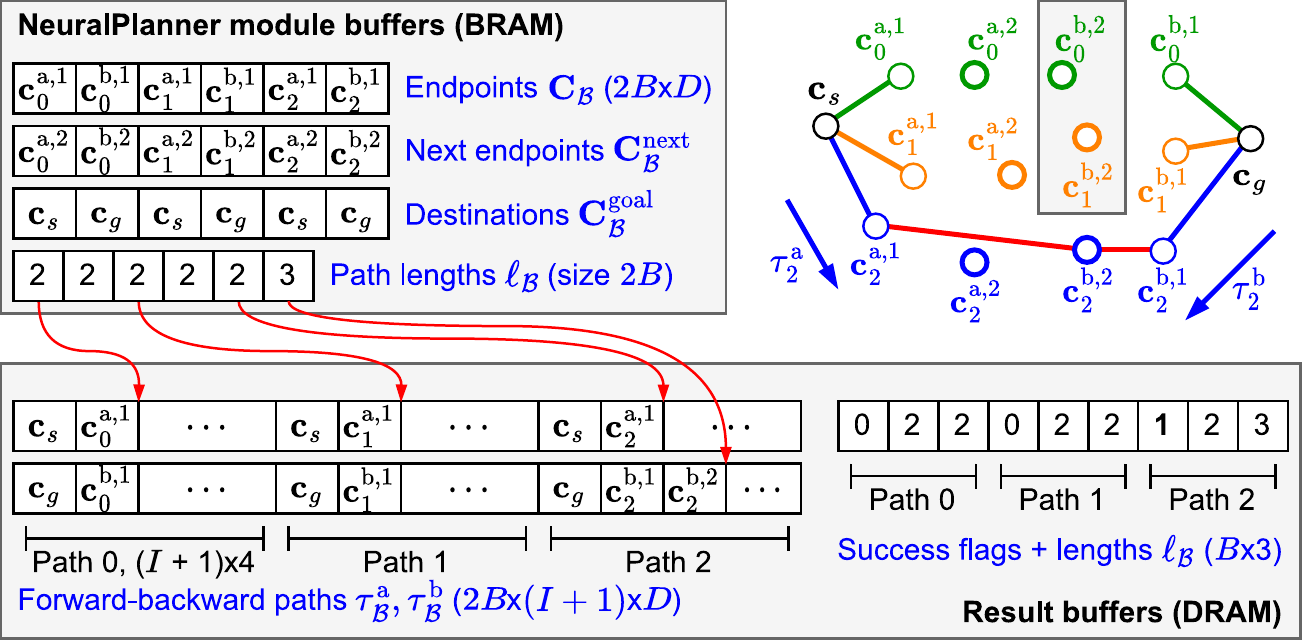}
  \caption{Buffer formats for NeuralPlanner module (batch size $B = 3$).
  NeuralPlanner updates the path endpoints $\mathbf{C}_\mathcal{B}, \mathbf{C}_\mathcal{B}^\mathrm{next}$, path lengths $\ell_\mathcal{B}$, as well as the DRAM buffers for forward-backward paths $\tau_\mathcal{B}^\mathrm{a} = \left[ \tau_0^\mathrm{a}, \tau_1^\mathrm{a}, \tau_2^\mathrm{a} \right], \tau_\mathcal{B}^\mathrm{b} = \left[ \tau_0^\mathrm{b}, \tau_1^\mathrm{b}, \tau_2^\mathrm{b} \right]$ at each iteration.
  The path lengths stored on-chip are used as pointers to the DRAM path buffers (\textcolor{red}{red arrows}).
  In iteration $i = 2$, the third path pair $\tau_2^\mathrm{a}, \tau_2^\mathrm{b}$ is found to be connectable.
  The module writes the success flags and path lengths $\ell_\mathcal{B}$ to the DRAM buffer and completes the task.}
  \label{fig:module-planner-buffers}
\end{figure}

As shown in Figs. \ref{fig:module-planner}-\ref{fig:module-planner-buffers}, the module interacts with submodules and manages several buffers to perform the bidirectional planning (Alg. \ref{alg:p3net-func-batch-plan}).
To start planning, the module first (1) reads the task configurations such as start-goal points $\mathbf{c}_s, \mathbf{c}_g$ from DRAM, as well as algorithmic parameters (e.g., the number of obstacles $N^\mathrm{obs}$, maximum iterations $I$, collision check step $\delta$, etc.) from control registers.
It then (2) initializes BRAM buffers (Fig. \ref{fig:module-planner-buffers} (top)) for the current endpoints, destinations, next waypoints $\mathbf{C}_\mathcal{B}, \mathbf{C}_\mathcal{B}^\mathrm{goal}, \mathbf{C}_\mathcal{B}^\mathrm{next} \in \mathbb{R}^{2B \times D}$ ($D = 2, 3$) and path lengths $\ell_\mathcal{B} \in \mathbb{N}^{2B}$, as well as the result buffers on DRAM (Alg. \ref{alg:p3net-func-batch-plan}, lines \ref{alg:p3net-np-init-endpoints}-\ref{alg:p3net-np-init-lengths}).
These result buffers store the entire forward-backward paths $\tau_\mathcal{B}^\mathrm{a}, \tau_\mathcal{B}^\mathrm{b}$ along with their lengths and success flags in a format depicted in Fig. \ref{fig:module-planner-buffers} (bottom).
Its size is bound by the maximum iterations $I$.
The on-chip length buffer $\ell_\mathcal{B}$ serves as a pointer to the end of path data (Fig. \ref{fig:module-planner-buffers}, \textcolor{red}{red} arrows).

The module proceeds to alternate between {\NewPNet} inference (line \ref{alg:p3net-np-pnet}) and collision checking (lines \ref{alg:p3net-np-check-begin}-\ref{alg:p3net-np-check-end}); it (3) writes {\NewPNet} inputs $\mathbf{C}_\mathcal{B}, \mathbf{C}_\mathcal{B}^\mathrm{goal}$ and computes the next waypoints $\mathbf{C}_\mathcal{B}^\mathrm{next}$ (Fig. \ref{fig:module-planner} (left bottom)).
The buffers for inference are implemented with BRAM or URAM.
Using the current and next endpoints ($\mathbf{c}_j^{\mathrm{a}, i}, \mathbf{c}_j^{\mathrm{b}, i} \in \mathbf{C}_\mathcal{B}$, $\mathbf{c}_j^{\mathrm{a}, i + 1}, \mathbf{c}_j^{\mathrm{b}, i + 1} \in \mathbf{C}_\mathcal{B}^\mathrm{next}$), the module (4) attempts to connect each path pair $\tau_j^\mathrm{a}, \tau_j^\mathrm{b}$ ($j \in \left[ 1, B \right)$) by the collision checking (Fig. \ref{fig:p3net-batched-planning}).
If the path connection succeeds, it transfers the success flag and path length ($\ell_j^\mathrm{a}, \ell_j^\mathrm{b} \in \ell_\mathcal{B}$) to the DRAM result buffer and completes the task.
The module appends new waypoints $\mathbf{C}_\mathcal{B}^\mathrm{next}$ to the DRAM result buffers, increments path lengths $\ell_\mathcal{B}$ accordingly, and updates the endpoint buffer $\mathbf{C}_\mathcal{B}$ with $\mathbf{C}_\mathcal{B}^\mathrm{next}$ for the next iteration.

As discussed in Sec. \ref{sec:impl-neural-planner-collision-check}, the collision check of a line segment ($\mathbf{c}^\mathrm{start}, \mathbf{c}^\mathrm{end}$) is cast as checks on the interpolated discrete points $\{ \mathbf{c}_i \}$.
The module reads a chunk of bounding boxes $\{ (\mathbf{c}_{i, \min}^\mathrm{obs}, \mathbf{c}_{i, \max}^\mathrm{obs}) \}$ into an on-chip obstacle buffer of size $(N_C^\mathrm{obs}, D)$ ($N_C^\mathrm{obs} = 64$), and tests each midpoint $\mathbf{c}_i$ with multiple obstacles in parallel using an array of $\mathrm{Check}$ submodules (Fig. \ref{fig:module-planner} (right bottom)).


\subsection{CollisionChecker Module} \label{sec:impl-collision-checker}
Since collision checking is performed throughout {\ProposalName}, it is implemented in a dedicated \textbf{CollisionChecker} module for further speedup.
It is in charge of testing the path $\tau = \left\{ \mathbf{c}_0, \ldots, \mathbf{c}_T \right\}$ by repeating the process described in Sec. \ref{sec:impl-neural-planner-collision-check} for each edge ($\mathbf{c}_i, \mathbf{c}_{i + 1}$).
The path $\tau$ and obstacle bounding boxes $\{ (\mathbf{c}_{i, \min}^\mathrm{obs}, \mathbf{c}_{i, \max}^\mathrm{obs}) \}$ are stored on the separate DRAM buffers, which are successively transferred to the on-chip buffers of size $(T_C, D)$ and $(N_C^\mathrm{obs}, D)$ ($T_C = 64, N_C^\mathrm{obs} = 64$).
The module checks the collision between obstacles and interpolated points on the edge $(\mathbf{c}_i, \mathbf{c}_{i + 1})$, and the result (1 if $\tau$ collides with any obstacle) is written to the control register.


\subsection{Board-level Implementation of {\IPName}} \label{sec:impl-board-level}
\begin{figure}
  \centering
  \includegraphics[keepaspectratio, width=0.65\linewidth]{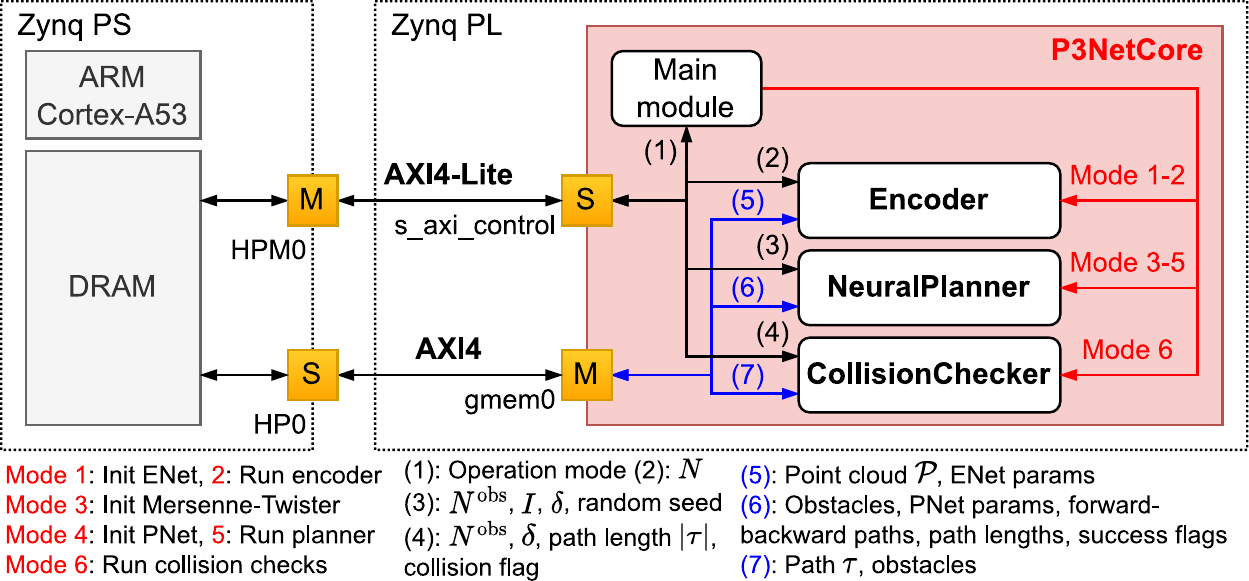}
  \caption{Board-level implementation of {\IPName}.}
  \label{fig:board-level-impl}
\end{figure}

\begin{figure}
  \centering
  \includegraphics[keepaspectratio, width=0.4\linewidth]{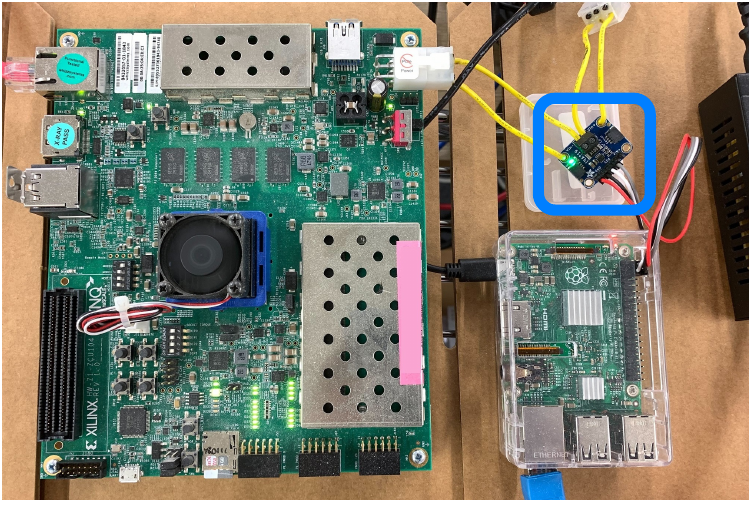}
  \caption{Xilinx ZCU104 Evaluation Kit with Texas Instruments INA219 power monitor (\textcolor{ProcessBlue}{skyblue}) and Raspberry Pi 3 Model B (bottom right).}
  \label{fig:xilinx-zcu104}
\end{figure}

Fig. \ref{fig:board-level-impl} shows the board-level implementation for Xilinx Zynq UltraScale+ MPSoC devices.
{\IPName} is implemented on the PL (Programmable Logic) part and communicates with the PS (Processing System) part via two AXI interfaces.
One is an AXI4-Lite subordinate interface with a 32-bit data bus connected to the High-Performance Master (HPM0) port, which allows PS part to access control registers.
The other is an AXI master interface with a 128-bit data bus connected to the High-Performance (HP0) port, through which the IP core transfers algorithm inputs/outputs in bursts (four 32-bit words per clock)\footnote{To ensure the data is 128-bit aligned, the DRAM buffer sizes are rounded up to the nearest multiple of four when necessary, e.g., the point cloud buffer is of size $(N, 4)$ instead of $(N, 2)$ or $(N, 3)$}.
We used Xilinx Vitis HLS 2022.1 to develop the IP core, and Vivado 2022.1 for synthesis and place-and-route.
Two variants of {\IPName} were created for 2D and 3D planning tasks.
The target FPGA SoC is a Xilinx ZCU104 Evaluation Kit (Fig. \ref{fig:xilinx-zcu104}, Table \ref{tbl:machines}).
The clock frequency of the board is set to 200MHz.



To preserve the precision of {\NewPNet} outputs (i.e., waypoint coordinates), Encoder and NeuralPlanner modules employ a 32-bit fixed-point with 16.16 format (i.e., a 16-bit integer part and 16-bit fraction part) for layer outputs, and 24-bit fixed-point with 8.16 format for model parameters.
Besides, collision checking is performed using the 32-bit floating-point, taking into account that the interval $\delta$ (Sec. \ref{sec:impl-neural-planner-collision-check}) is set sufficiently small to prevent false negatives.


As shown in Fig. \ref{fig:board-level-impl}, {\IPName} has six operation modes, namely, \textbf{Init\{ENet, MT, PNet\}} and \textbf{Run\{Encoder, Planner, CollisionChecker\}}.
When the IP core is called from the CPU, it first reads an operation mode from the control register and performs the corresponding task.
\textbf{Init\{E, P\}Net} is responsible for transferring the parameters of {\NewEPNet} from DRAM to the dedicated on-chip buffers via an AXI master port.
When \textbf{InitMT} is specified, the main module reads a 32-bit random seed from the register and initializes the internal states (624 words) of Mersenne-Twister for dropout layers (Sec. \ref{sec:impl-neural-planner-inference}).
\textbf{Run\{Encoder, Planner, CollisionChecker\}} triggers the respective module (Sec. \ref{sec:impl-encoder}-\ref{sec:impl-collision-checker}).
The transferred data is summarized in Fig. \ref{fig:board-level-impl} (bottom).

\section{Evaluation} \label{sec:eval}
This section evaluates the proposed {\ProposalName} to demonstrate the improvements on success rate, speed, quality of solutions, and power efficiency.
For convenience, {\ProposalName} accelerated by the proposed IP core is referred to as \textbf{{\ProposalNameFPGA}}.

\subsection{Experimental Setup} \label{sec:eval-setup}
MPNet and the famous sampling-based methods, RRT*~\cite{SertacKaraman11}, Informed-RRT* (IRRT*)~\cite{JonathamGammell14}, BIT*~\cite{JonathamGammell15}, and ABIT*~\cite{MarlinPStrub20A} were used as a baseline.
All methods including {\ProposalName} were implemented in Python using NumPy and PyTorch.
The implementation of RRT*, Informed-RRT*, and BIT* is based on the open-source code~\cite{HuimingZhouPathPlanning}\footnote{For a fair performance comparison, we replaced a brute-force linear search with an efficient K-D tree-based search using Nanoflann library (v1.4.3)~\cite{JoseLuisBlanco14}. In BIT* planner, we used a priority queue to pick the next edge to process, which avoids searching the entire edge queue, as mentioned in the original paper~\cite{JonathamGammell15}.}.
We implemented an ABIT* planner following the algorithm in the paper~\cite{MarlinPStrub20A}.
For MPNet, we used the code from the authors~\cite{AhmedQureshi21} as a reference; MPNet path planner and the other necessary codes for training and testing were rewritten from scratch.


The experiments were conducted on a workstation, Nvidia Jetson \{Nano, Xavier NX\}, and Xilinx ZCU104; their specifications and environments are summarized in Table \ref{tbl:machines}.
For ZCU104, we built PyTorch from source with \texttt{-O3} optimization level and ARM Neon intrinsics enabled (\texttt{-D\_\_NEON\_\_}) to fully take advantage of the multicore architecture.

Following the MPNet paper~\cite{AhmedQureshi21}, we trained encoding and planning networks in an end-to-end supervised manner, which is outlined as follows.
Given a training sample $(\mathcal{P}, \mathbf{c}_t^*, \mathbf{c}^\mathrm{goal}, \mathbf{c}_{t + 1}^*)$, where $\mathbf{c}_t^*, \mathbf{c}_{t + 1}^*$ denote a pair of waypoints on a ground-truth path\footnote{The ground-truth path is obtained by running an RRT* planner with a large number of iterations.}, the encoder first extracts a global feature $\boldsymbol{\phi}(\mathcal{P})$, and then the planning network estimates the next waypoint $\mathbf{c}_{t + 1}$.
The squared Euclidean distance $\| \mathbf{c}_{t + 1} - \mathbf{c}_{t + 1}^* \|^2$ is used as a loss function, so that two models jointly learn to mimic the behavior of the planner used for dataset generation.
We used an Adam optimizer with a learning rate of $10^{-3}$ and coefficients of $\beta_1 = 0.9, \beta_2 = 0.999$.
We set the number of epochs to 200 and 50 when training MPNet and {\ProposalName}, and the batch size to 8192 for MPNet (2D), 1024 for MPNet (3D), and 128 for {\ProposalName} models, respectively.
The number of iterations $I$ in $\mathrm{NeuralPlanner}$($\mathrm{Ex}$) (Algs. \ref{alg:mpnet}, \ref{alg:p3net-func-batch-plan}) is fixed to 50.


RRT* and Informed-RRT* were configured with a maximum step size of 1.0 and a goal bias of $0.05$.
BIT* and ABIT* were executed with a batch size of 100 and Euclidean distance heuristic.
As done in \cite{MarlinPStrub20A}, ABIT* searched the same batch twice with two different inflation factors $\varepsilon_\mathrm{infl} = 10^6, 1 + \frac{10}{q}$, and the truncation factor $\varepsilon_\mathrm{trunc}$ was set to $1 + \frac{5}{q}$, where $q$ is a total number of nodes.

\begin{table}[h]
  \small
  \centering
  \caption{Evaluation Environments}
  \label{tbl:machines}
  \begin{tabular}{l|ll} \hline
    & Workstation & Nvidia Jetson Nano \\ \hline
    \multirow{2}{*}{CPU} & Intel Xeon W-2235 & ARM Cortex-A57 \\
    & @3.8GHz, 12C & @1.43GHz, 4C \\ \hline
    GPU & Nvidia GeForce RTX 3090 & 128-core Nvidia Maxwell \\ \hline
    DRAM & 64GB (DDR4) & 4GB (DDR4) \\ \hline
    \multirow{2}{*}{OS} & \multirow{2}{*}{Ubuntu 20.04.6} & Nvidia JetPack 4.6.3 \\
    & & (based on Ubuntu 18.04) \\ \hline
    Python & 3.8.2 & 3.6.15 \\
    PyTorch & 1.11.0 (w/ CUDA 11.3) & 1.10.0 (w/ CUDA 10.2) \\ \hline \hline

    & Nvidia Jetson Xavier NX & Xilinx ZCU104 Evaluation Kit \\ \hline
    \multirow{2}{*}{CPU} & Nvidia Carmel ARM v8.2 & ARM Cortex-A53 \\
    & @1.4GHz, 6C & @1.2GHz, 4C \\ \hline
    GPU & 384-core Nvidia Volta & -- \\ \hline
    FPGA & -- & XCZU7EV-2FFVC1156 \\ \hline
    DRAM & 8GB (DDR4) & 2GB (DDR3) \\ \hline
    \multirow{2}{*}{OS} & Nvidia JetPack 5.1 & Pynq Linux v2.7 \\
    & (based on Ubuntu 20.04) & (based on Ubuntu 20.04) \\ \hline
    Python & 3.8.2 & 3.8.2 \\
    PyTorch & 1.14.0a0 (w/ CUDA 11.4) & 1.10.2 \\ \hline
  \end{tabular}
\end{table}



\subsection{Path Planning Datasets} \label{sec:eval-dataset}
For evaluation, we used a publicly-available dataset for 2D/3D path planning provided by the MPNet authors~\cite{AhmedQureshi21}, referred to as \textbf{MPNet2D}/\textbf{3D}\footnote{Originally called Simple2D and Complex3D in the MPNet paper.}.
It is split into one training and two testing sets (\textbf{Seen}, \textbf{Unseen}); the former contains 100 workspaces, each of which has a point cloud $\mathcal{P}$ representing obstacles, and 4000 planning tasks with randomly generated start-goal points and their respective ground-truth paths.
\textbf{Seen} set contains the same 100 workspaces as the training set, but with each having 200 new planning tasks; \textbf{Unseen} set comes with ten new workspaces not observed during training, each of which has 2000 planning tasks.

Each workspace is a square (or cube) of size 40 containing randomly-placed seven square obstacles of size 5 (or ten cuboid obstacles with a side length of 5 and 10).
Note that, trivial tasks in the testing sets are excluded, where start-goal pairs can be connected by straight lines and obstacle avoidance is not required.
We only used the first 20/200 tasks for each workspace in \textbf{Seen}/\textbf{Unseen} dataset.
As a result, the total number of planning tasks is 945/892 and 740/756 in MPNet2D (Seen/Unseen) and MPNet3D (Seen/Unseen) datasets, respectively.
Both MPNet and {\ProposalName} models were trained with MPNet2D/3D training sets.

In addition, we generated \textbf{{\ProposalName}2D}/\textbf{3D} dataset for testing (Figs. \ref{fig:result-p3net2d}-\ref{fig:result-p3net3d}), which contains 100 workspaces, with 20 planning tasks for each.
Compared to MPNet2D/3D, the number of obstacles is doubled to simulate more challenging tasks.

\subsection{Planning Success Rate} \label{sec:eval-success-rate}
First, the tradeoff between planning success rate and the average computation time per task is evaluated on the workstation with GPU acceleration.

\subsubsection{Comparison of Encoding and Planning Networks} \label{sec:eval-success-rate-models}
MPNet is executed under three combinations of models, i.e., {\EPNet} (original), {\NewENet}-PNet (\textbf{ELite}), and {\NewEPNet} (\textbf{EPLite}), with a varying number of replan iterations $I_\mathrm{Replan} = \{10, 20, 50, 100\}$.
For {\ProposalName}, the refinement step is not performed ($I_\mathrm{Refine} = 0$).
The results on MPNet and {\ProposalName} test datasets are shown in Fig. \ref{fig:mpnet-p3net-success-rate-time} (left).

In the 2D case (1st/3rd row), replacing {\EPNet} with {\NewEPNet} yields substantially higher success rate and even faster computation time, while reducing the parameters by 32.32x (Sec. \ref{sec:method-improved-pnet}).
For $I_\mathrm{Replan} = 10$, MPNet with ELite setting is 15.58\% (69.17/84.75\%) and 23.75\% (45.15/68.90\%) more successful than the original setting on MPNet (Unseen) and {\ProposalName} datasets, respectively; {\NewPNet} further improves the success rate by 3.48\% (84.75/88.23\%) and 4.45\% (68.90/73.35\%).
MPNet (EPLite) is 1.40x (0.067/0.048s) and 1.12x (0.131/0.117s) faster than MPNet (original), indicating that the proposed models help MPNet algorithm to quickly find a solution in a less number of replan attempts.
This empirically validates the discussion in Sec. \ref{sec:method-improved-enet} that the shallower PNet is sufficient since the PointNet encoder produces more robust and informative features.
Considering the success rate improvements (19.06/28.20\%) in these two datasets, the proposed models offer greater performance advantages in more difficult problem settings.


In the 3D case (Fig. \ref{fig:mpnet-p3net-success-rate-time} (2nd/4th row, left)), MPNet (EPLite) maintains the same level of performance as MPNet (original), while achieving 5.43x parameter reduction (Sec. \ref{sec:method-improved-pnet}).
{\NewENet} improves the success rate by 1.46\% (89.82/91.27\%) and 3.70\% (79.35/83.05\%), whereas {\NewPNet} slightly lowers it by 0.40\% (91.27/90.87\%) and 3.95\% (83.05/79.10\%) on MPNet (Unseen) and {\ProposalName} datasets.
This performance loss is compensated by the {\ProposalName} planner.
Comparing the results from MPNet \textcolor{darkgray}{Seen}/\textcolor{blue}{Unseen} datasets (dashed/solid lines), the difference in the success rate is at most 3.46\%, which confirms that our proposed models generalize to workspaces that are not observed during training.





\subsubsection{Comparison of {\ProposalName} with MPNet} \label{sec:eval-success-rate-mpnet}
Fig. \ref{fig:mpnet-p3net-success-rate-time} (left) also highlights the advantage of {\ProposalName} ($B = \{1, 2, 4, 8\}, I_\mathrm{Replan} = \{10, 20, 50, 100\}$).
Though MPNet exhibits gradual improvement in success rate with increasing $I_\mathrm{Replan}$, {\ProposalName} consistently outperforms MPNet and achieves nearly 100\% success rate\footnote{Note that the success rate of {\ProposalName} ($B = 1$) surpasses that of MPNet (EPLite), since the former performs more collision checks to connect a pair of forward-backward paths in each iteration (Alg. \ref{alg:mpnet}, line \ref{alg:mpnet-np-check-connectable} and Alg. \ref{alg:p3net-func-batch-plan}, lines \ref{alg:p3net-np-check-begin}-\ref{alg:p3net-np-check-end}).}.
In the 2D case (1st/3rd row), {\ProposalName} ($B, I_\mathrm{Replan} = 8, 100$) is 2.80\% (96.30/99.10\%) and 9.85\% (87.60/97.45\%) more likely to find a solution in 2.20x (0.101/0.046s) and 1.55x (0.326/0.210s) shorter time than MPNet (EPLite, $I_\mathrm{Replan} = 100$) on MPNet (Unseen) and {\ProposalName} datasets.
While MPNet shows a noticeable drop in success rate (96.30/87.69\%) when tested on {\ProposalName} dataset, {\ProposalName} only shows a 1.65\% drop (99.10/97.45\%) and maintains the high success rate in a more challenging dataset.
In the 3D case (2nd/4th row), it is 2.91\% (96.69/99.60\%) and 8.00\% (91.75/99.75\%) better while spending 2.70x (0.081/0.030s) and 1.02x (0.277/0.272s) less time.

Notably, increasing the batch size $B$ improves both success rate and speed, which clearly indicates the effectiveness of the batch planning strategy (Sec. \ref{sec:method-batch-planning}).
On {\ProposalName}2D dataset (3rd row), {\ProposalName} with $B, I_\mathrm{Replan} = 8, 10$ is 5.25\% more successful and 1.88x faster than with $B = 1$ (79.40/84.65\%, 0.128/0.068s).
The number of initial planning attempts $I_\mathrm{Init}$ also affects the performance; increasing it from 1 to 5 yields a 2.85\% better success rate on {\ProposalName}2D dataset ($I_\mathrm{Replan} = 10$).
Table \ref{tbl:success-rate-p3netip} compares the success rate of {\ProposalNameFPGA} and {\ProposalName}; despite the use of fixed-point arithmetic and a simple pseudorandom generator, {\ProposalNameFPGA} attains a similar performance.

\subsubsection{Comparison with Sampling-based Methods} \label{sec:eval-success-rate-sampling}
\begin{figure}[h]
  \centering
  \includegraphics[keepaspectratio, width=0.8\linewidth]{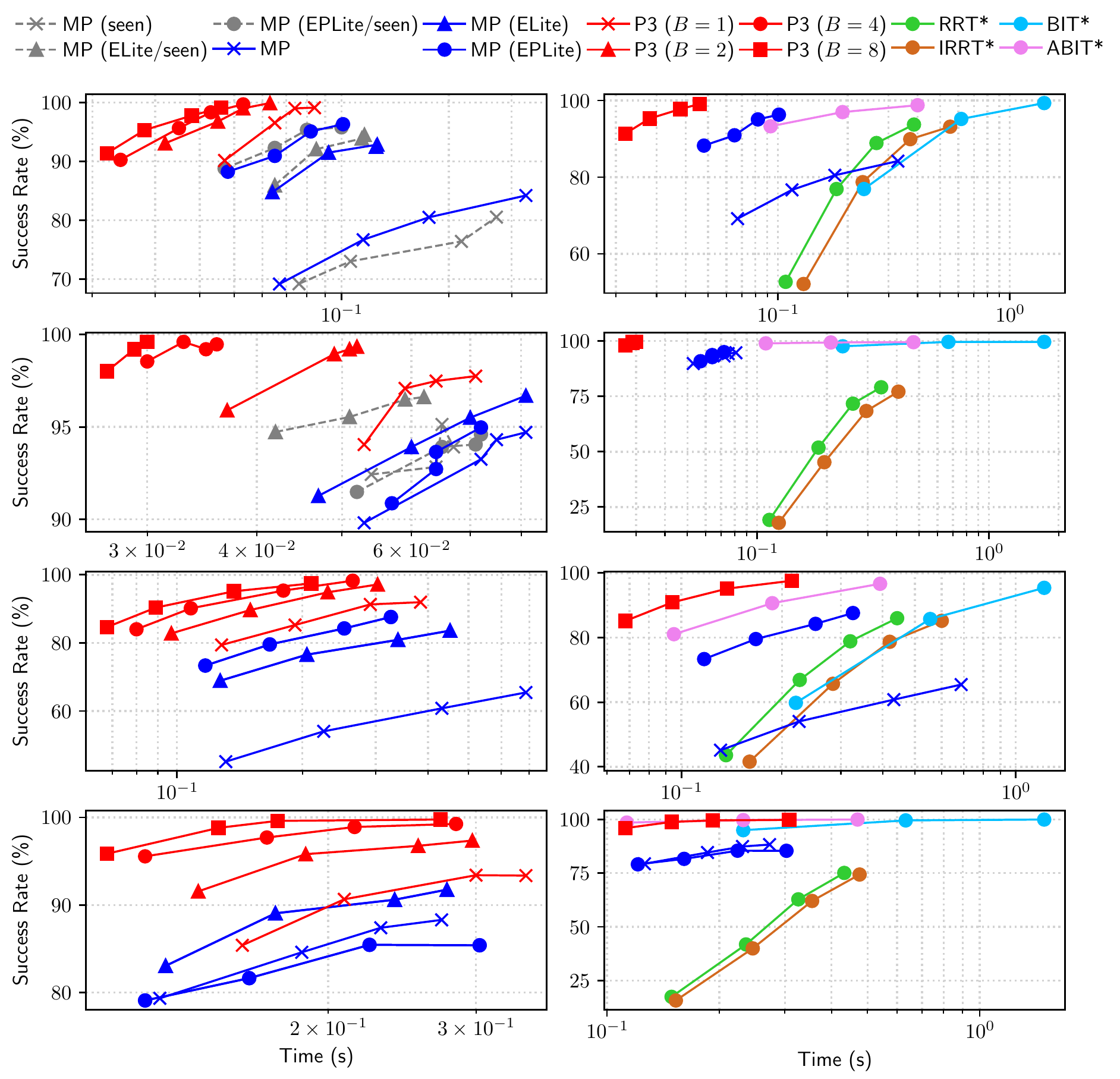}
  \caption{Success rate and computation time tradeoffs (MPNet\{2D, 3D\}, {\ProposalName}\{2D, 3D\} dataset from top to bottom, measured on the workstation with GPU acceleration). Upper left is better.}
  \label{fig:mpnet-p3net-success-rate-time}
\end{figure}

Fig. \ref{fig:mpnet-p3net-success-rate-time} (right) plots the results from sampling-based methods.
The number of iterations is set to $\{200, 300, 400, 500\}$ for RRT*/IRRT*, and $\{50, 100, 200\}$ for BIT*/ABIT*.
As expected, sampling-based methods exhibit a higher success rate with increasing iterations; they are more likely to find a solution as they place more random nodes inside a workspace and build a denser tree.
{\ProposalName} achieves a success rate comparable to ABIT*, and outperforms the other methods.
On {\ProposalName}2D dataset, {\ProposalName} ($B, I_\mathrm{Replan} = 8, 100$) plans a path in an average of 0.214s with 97.60\% success rate, which is 1.84/5.69x faster and 0.95/2.20\% better than ABIT*/BIT* (200 iterations).



\begin{table}[h]
  \small
  \centering
  \caption{Success rate of {\ProposalNameFPGA} and {\ProposalName} (w/ and w/o {\IPName})}
  \label{tbl:success-rate-p3netip}
  \begin{tabular}{rrrr|rrrr} \hline
    \multicolumn{4}{c|}{MPNet (2D) dataset} &
    \multicolumn{4}{c}{MPNet (3D) dataset} \\ \hline
    $B$ & $I_\mathrm{Replan}$ & \% (w/) & \% (w/o) &
    $B$ & $I_\mathrm{Replan}$ & \% (w/) & \% (w/o) \\ \hline
    4 & 10 & 91.70 & 91.26 &
    4 & 10 & 97.88 & 97.88 \\
    4 & 20 & 95.18 & 94.84 &
    4 & 20 & 99.07 & 99.47 \\
    4 & 50 & 98.54 & 98.54 &
    4 & 50 & 99.60 & 99.60 \\
    4 & 100 & 99.78 & 99.66 &
    4 & 100 & 99.60 & 99.60 \\ \hline \hline
    \multicolumn{4}{c|}{{\ProposalName} (2D) dataset} &
    \multicolumn{4}{c}{{\ProposalName} (3D) dataset} \\ \hline
    $B$ & $I_\mathrm{Replan}$ & \% (w/) & \% (w/o) &
    $B$ & $I_\mathrm{Replan}$ & \% (w/) & \% (w/o) \\ \hline
    4 & 10 & 84.75 & 84.25 &
    4 & 10 & 94.95 & 95.60 \\
    4 & 20 & 91.40 & 91.00 &
    4 & 20 & 98.90 & 98.90 \\
    4 & 50 & 95.45 & 95.45 &
    4 & 50 & 99.60 & 99.75 \\
    4 & 100 & 98.25 & 97.80 &
    4 & 100 & 99.80 & 100.0 \\ \hline
  \end{tabular}
\end{table}

\subsection{Computation Time} \label{sec:eval-computation-time}
\begin{figure}[h]
  \centering
  \includegraphics[keepaspectratio, width=0.65\linewidth]{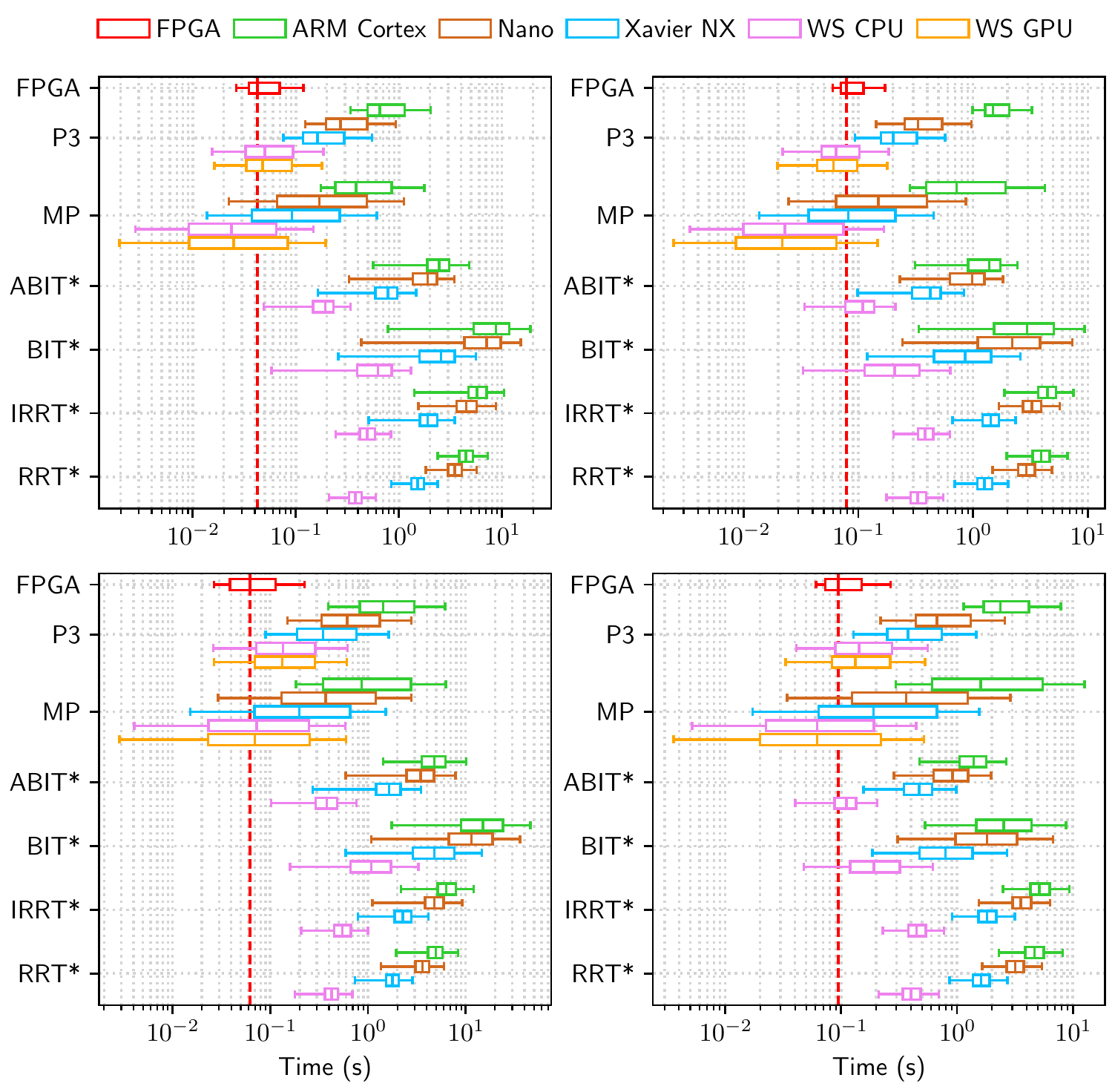}
  \caption{Computation time distribution (top: MPNet, bottom: {\ProposalName} dataset, left: 2D, right: 3D dataset).
  Hyperparameters are as follows.
  \underline{{\ProposalName}2D}: $(B, I_\mathrm{Init}, I_\mathrm{Replan}, I_\mathrm{Refine}) = (4, 5, 50, 5)$,
  \underline{{\ProposalName}3D}: $(B, I_\mathrm{Init}, I_\mathrm{Replan}, I_\mathrm{Refine}) = (4, 5, 20, 5)$,
  \underline{MPNet2D/3D}: {\NewEPNet}, $I_\mathrm{Replan} = 100$,
  \underline{ABIT*/BIT* (2D)}: 100/200 iterations for MPNet/{\ProposalName} dataset,
  \underline{ABIT*/BIT* (3D)}: 50 iterations,
  \underline{IRRT*/RRT* (2D/3D)}: 500 iterations.
  The \textcolor{red}{red dashed line} is a median time of {\ProposalNameFPGA}.}
  \label{fig:mpnet-p3net-time}
\end{figure}

Fig. \ref{fig:mpnet-p3net-time} visualizes the distribution of computation time measured on the workstation and SoC devices (Table \ref{tbl:machines}).
The sampling-based methods were run on the CPU.
On Nvidia Jetson, MPNet and {\ProposalName} were executed with GPU.
WS \{CPU, GPU\} refers to the workstation with and without GPU acceleration.
On the basis of results from Sec. \ref{sec:eval-success-rate}, hyperparameters of the planners were selected to achieve similar success rates.
For a fair comparison, the PS--PL data transfer overhead is included in {\ProposalNameFPGA}.

As expected, {\ProposalNameFPGA} is faster than the other planners on ZCU104 and Jetson (\textcolor{Green}{green}, \textcolor{Bittersweet}{brown}, \textcolor{ProcessBlue}{skyblue}) in most cases, and its median computation time is below 0.1s.
In the 2D case (Fig. \ref{fig:mpnet-p3net-time} (left)), it even outperforms sampling-based methods on the WS CPU (\textcolor{CarnationPink}{pink}), and is comparable to MPNet/{\ProposalName} on the WS GPU (\textcolor{Orange}{orange}).
On {\ProposalName}2D dataset, {\ProposalNameFPGA} takes 0.062s in the median to solve a task, which is 2.15x, 1.13x, 6.11x, and 17.52x faster than {\ProposalName} (GPU), MPNet (GPU), ABIT*, and BIT* on the workstation.
{\ProposalName} offers more performance advantages in a more challenging dataset.
On the ZCU104 and {\ProposalName}/MPNet2D dataset, {\ProposalNameFPGA} is 15.57x/9.17x faster than MPNet (CPU).



In the 3D case (Fig. \ref{fig:mpnet-p3net-time} (right)), while MPNet on the workstation looks faster than {\ProposalNameFPGA}, it only solves easy planning tasks which require less replan trials, and the result shows around 10\% lower success rate.
We observe a significant reduction of the variance in computation time.
On the ZCU104 and {\ProposalName} dataset (Fig. \ref{fig:mpnet-p3net-time}, bottom right), {\ProposalName} solves a task in 7.623$\pm$19.987s on average, which is improved to 4.651$\pm$8.909s and 0.145$\pm$0.170s in {\ProposalName} (CPU, FPGA).
As seen from the results of WS \{CPU, GPU\}, GPU acceleration of the DNN inference does not contribute to the overall performance improvement; since MPNet/{\ProposalName} alternates between collision checks (CPU) and PNet inference (GPU), the frequent CPU--GPU data transfer undermines the performance gain obtained by GPU.





\subsubsection{Path Planning Speedup} \label{sec:eval-computation-time-speedup}
\begin{figure}[h]
  \centering
  \includegraphics[keepaspectratio, width=0.7\linewidth]{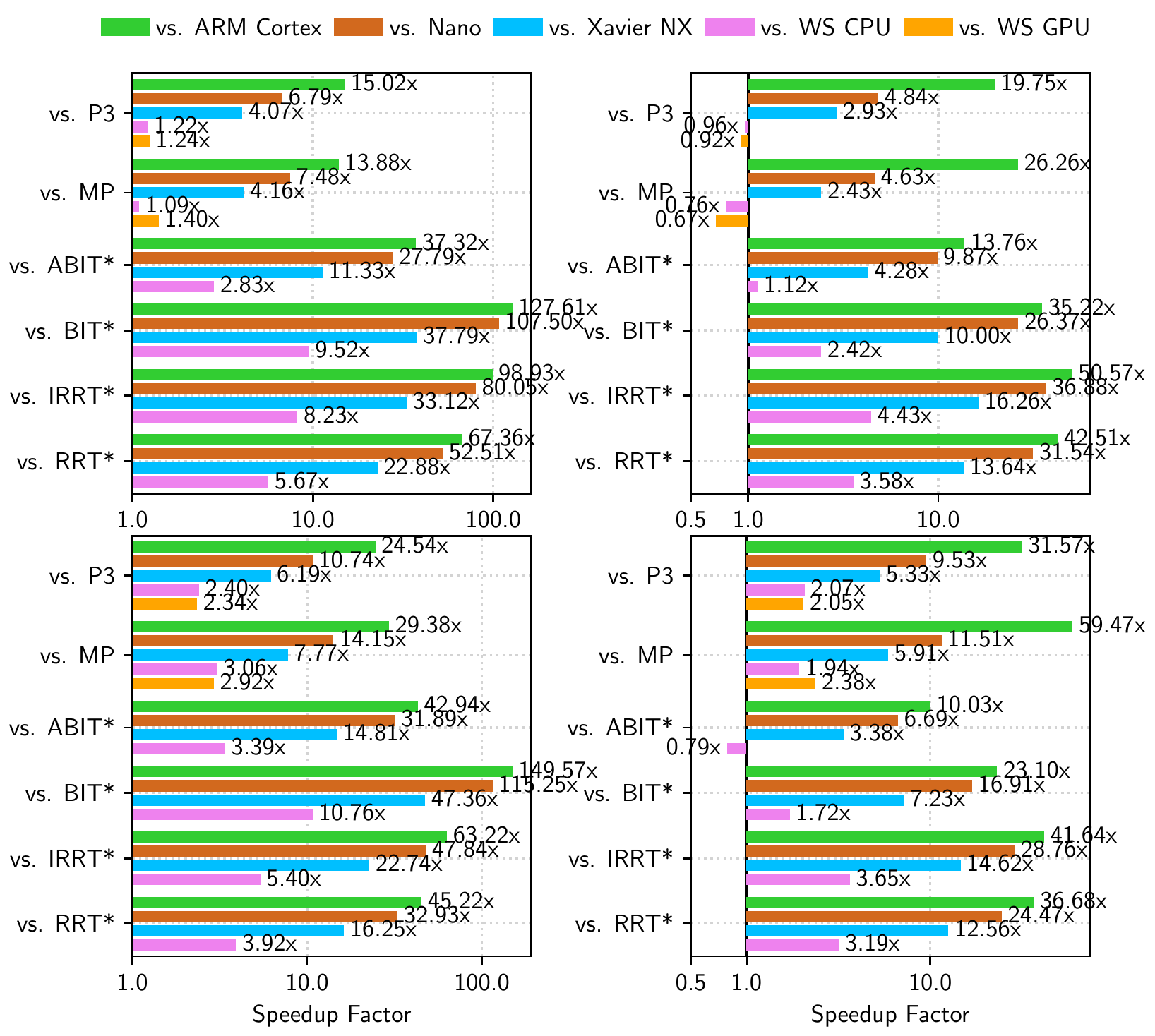}
  \caption{Average speedup factors (top: MPNet, bottom: {\ProposalName}, left: 2D, right: 3D dataset).
  Hyperparameter settings are the same as in Fig. \ref{fig:mpnet-p3net-time}.}
  \label{fig:mpnet-p3net-time-compare}
\end{figure}

Fig. \ref{fig:mpnet-p3net-time-compare} shows the performance gain of {\ProposalNameFPGA} over the other planners on the workstation and SoC devices.
In the 2D case, {\ProposalNameFPGA} is the fastest among the methods considered.
On {\ProposalName} dataset (bottom left), it achieves 24.54--149.57x, 10.74--115.25x, 6.19--47.36x, and 2.34--10.76x speedups over the ZCU104, Jetson Nano, Jetson Xavier NX, and a workstation, respectively.
Offloading the entire planning algorithm (collision checks and neural planning in Alg. \ref{alg:p3net-func-batch-plan}) to the dedicated IP core eliminates unnecessary data transfers and brings more performance benefits than highend CPUs.
In this evaluation, we compared the sum of execution times for successful planning tasks to compute average speedup factors.
Unlike MPNet, {\ProposalName} also performs an extra refinement phase ($I_\mathrm{Refine} = 5$).
This means {\ProposalNameFPGA} completes more planning tasks in a shorter period of time than MPNet (e.g., 7.85\% higher success rate and 2.92x speedup than MPNet (WS GPU)).
Additionally, while {\ProposalNameFPGA} shows a 1.13x speedup over {\ProposalName} (WS GPU) in terms of median time (Fig. \ref{fig:mpnet-p3net-time} (bottom left)), the total execution time is reduced by 2.92x (Fig. \ref{fig:mpnet-p3net-time-compare} (bottom left)).
This implies {\ProposalName} solves challenging tasks much faster than MPNet, which is attributed to the improved algorithm and model architecture.
Also in the 3D case, {\ProposalNameFPGA} outperforms the other planners except a few cases.
On {\ProposalName}3D dataset (Fig. \ref{fig:mpnet-p3net-time-compare} (bottom right)), it provides 10.03--59.47x, 6.69--28.76x, 3.38--14.62x, and 0.79--3.65x speedups compared to the ZCU104, Jetson Nano, Jetson Xavier NX, and a workstation, respectively.





\subsubsection{Computation Time Breakdown} \label{sec:eval-computation-time-breakdown}
\begin{figure}[h]
  \centering
  \includegraphics[keepaspectratio, width=0.7\linewidth]{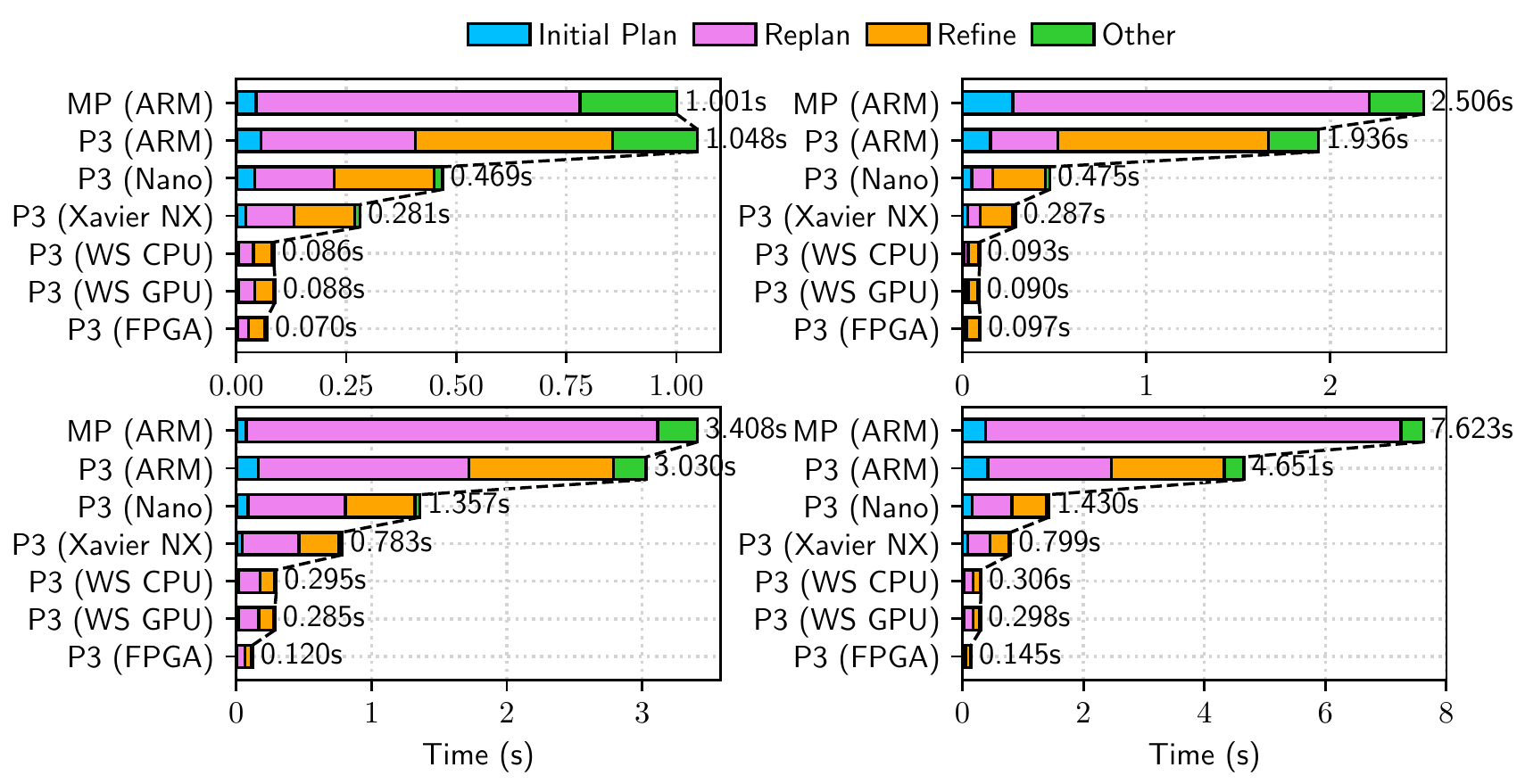}
  \caption{Computation time breakdown of {\ProposalNameFPGA} in comparison with MPNet and {\ProposalName} (top: MPNet, bottom: {\ProposalName}, left: 2D, right: 3D dataset).
  Hyperparameter settings are the same as in Fig. \ref{fig:mpnet-p3net-time}.}
  \label{fig:time-breakdown}
\end{figure}

The computation time breakdown of MPNet/{\ProposalName} is summarized in Fig. \ref{fig:time-breakdown}.
{\IPName} effectively reduces the execution time of all three phases.
On {\ProposalName}2D dataset (bottom left), the replanning phase (\textcolor{CarnationPink}{pink}) took 3.041s and accounted for 89.24\% of the entire execution time in MPNet, which is almost halved to 1.557s (51.37\%) in {\ProposalName} (CPU), and is brought down to only 0.062s (51.46\%) in {\ProposalNameFPGA}.
The saved time can be used to perform the additional refinement step (\textcolor{Orange}{orange}) and improve the quality of solutions.



\subsubsection{Speedup of Inference and Collision Checking} \label{sec:eval-computation-time-speedup2}
Table \ref{tbl:inference-collision-check-time} (top) lists the inference time of {\EPNet} and {\NewEPNet}, measured on the ZCU104 with and without the {\IPName}.
The data is the average of 50 runs.
As mentioned in Sec. \ref{sec:impl-encoder}, {\NewENet} basically involves $N$ times of forward passes to compute pointwise features, which increases the inference time by 3.78/1.57x than ENet2D/3D ($N = 1400, 2000$).
{\IPName} accelerates the inference by 45.47/45.34x and as a result achieves 12.01/28.88x faster feature extraction than ENet.
{\IPName} consistently attains around a 45x speedup on a wide range of $N$, owing to the combination of inter-layer pipelining (Fig. \ref{fig:module-encoder} (top)) and parallelization within each layer.
The inference time increases proportionally to $N$, which corresponds to the $\mathcal{O}(N)$ complexity of {\NewENet}, and {\IPName} yields a better performance gain with a larger $N$ (e.g., 45.47/47.13x in $N = 1400, 4096$, 2D).




{\NewPNet} has a 28.87/2.32x shorter inference time than PNet2D/3D, mainly due to the 32.58/2.35x parameter reduction (Sec. \ref{sec:method-improved-pnet}).
It does not grow linearly with the batch size $B$, indicating that the batch planning strategy in {\ProposalName} effectively improves success rates without incurring a significant additional overhead.
With {\IPName}, {\NewPNet} is sped up by 10.15--11.49/16.55--17.44x, resulting in an overall speedup of 280.45--326.25/38.27--40.48x over PNet.






Table \ref{tbl:inference-collision-check-time} (bottom) presents the computation time of collision checking on the ARM Cortex CPU and {\IPName}.
We conducted experiments with four random workspaces of size 40 containing different numbers of obstacles $N^\mathrm{obs}$, and 50 random start-goal pairs with a fixed distance of $d = 5.0, 20.0$ for each workspace.
The interval $\delta$ is set to 0.01.
{\IPName} gives a larger speedup with the increased problem complexity (larger $d$ or $N^\mathrm{obs}$).
It runs 2D/3D collision checking 10.49--55.19/31.92--221.45x (11.34--60.56/30.82--181.12x) faster in case of $d = 5.0$ and $20.0$, respectively.
The parallelization with an array of pipelined submodules (Sec. \ref{sec:impl-encoder-flow}) contributes to the two orders of magnitude performance improvement.
The result from {\IPName} does not show a linear increase with $N^\mathrm{obs}$, as {\IPName} terminates the checks as soon as any of midpoints between start-goal points is found to be in collision, and such early-exit is more likely to occur in a cluttered workspace with more obstacles.
A slight latency jump at $N^\mathrm{obs} = 128$ is due to the limited buffer size for obstacle positions and the increased data transfer from DRAM.





\begin{table}[h]
  \small
  \centering
  \caption{Latency for inference (top) and collision checking (bottom) on Xilinx ZCU104}
  \label{tbl:inference-collision-check-time}
  \begin{tabular}{l|rrr|rrr} \hline
    & \multicolumn{3}{c|}{2D} & \multicolumn{3}{c}{3D} \\ \hline
    Model & $N$ & CPU (ms) & \textbf{IP} (ms) & $N$ & CPU (ms) & \textbf{IP} (ms) \\ \hline
    ENet & 1400 & 43.25 & -- & 2000 & 146.40 & -- \\ \hline
    \multirow{3}{*}{\textbf{{\NewENet}}} & 1400 & 163.70 & 3.60 & 2000 & 229.85 & 5.07 \\
    & 2048 & 242.05 & 5.18 & 4096 & 478.06 & 10.15 \\
    & 4096 & 477.47 & 10.13 & 8192 & 950.91 & 20.11 \\ \hline
    Model & $B$ & CPU (ms) & \textbf{IP} (ms) & $B$ & CPU (ms) & \textbf{IP} (ms) \\ \hline
    \multirow{3}{*}{PNet} & 1 & 102.77 & -- & 1 & 103.22 & -- \\
    & 2 & 107.79 & -- & 2 & 108.31 & -- \\
    & 4 & 130.41 & -- & 4 & 130.68 & -- \\ \hline
    \multirow{3}{*}{\textbf{{\NewPNet}}} & 1 & 3.56 & 0.315 & 1 & 44.48 & 2.55 \\
    & 2 & 4.02 & 0.350 & 2 & 46.85 & 2.83 \\
    & 4 & 4.72 & 0.465 & 4 & 56.63 & 3.38 \\ \hline \hline
    & \multicolumn{3}{c|}{2D} & \multicolumn{3}{c}{3D} \\ \hline
    Dist. & $N^\mathrm{obs}$ & CPU (ms) & \textbf{IP} (ms) &
    $N^\mathrm{obs}$ & CPU (ms) & \textbf{IP} (ms) \\ \hline
    \multirow{4}{*}{5.0} & 16 & 3.02 & 0.288 & 16 & 3.29 & 0.290 \\
    & 32 & 5.23 & 0.290 & 32 & 5.75 & 0.291 \\
    & 64 & 9.65 & 0.284 & 64 & 10.68 & 0.291 \\
    & 128 & 18.49 & 0.335 & 128 & 20.59 & 0.340 \\ \hline
    \multirow{4}{*}{20.0} & 16 & 10.12 & 0.317 & 16 & 11.22 & 0.364 \\
    & 32 & 18.97 & 0.300 & 32 & 21.18 & 0.380 \\
    & 64 & 36.91 & 0.301 & 64 & 41.09 & 0.386 \\
    & 128 & 72.86 & 0.329 & 128 & 80.78 & 0.446 \\ \hline
  \end{tabular}
\end{table}

\subsection{Path Cost} \label{sec:eval-path-cost}
\begin{figure}[h]
  \centering
  \includegraphics[keepaspectratio, width=0.7\linewidth]{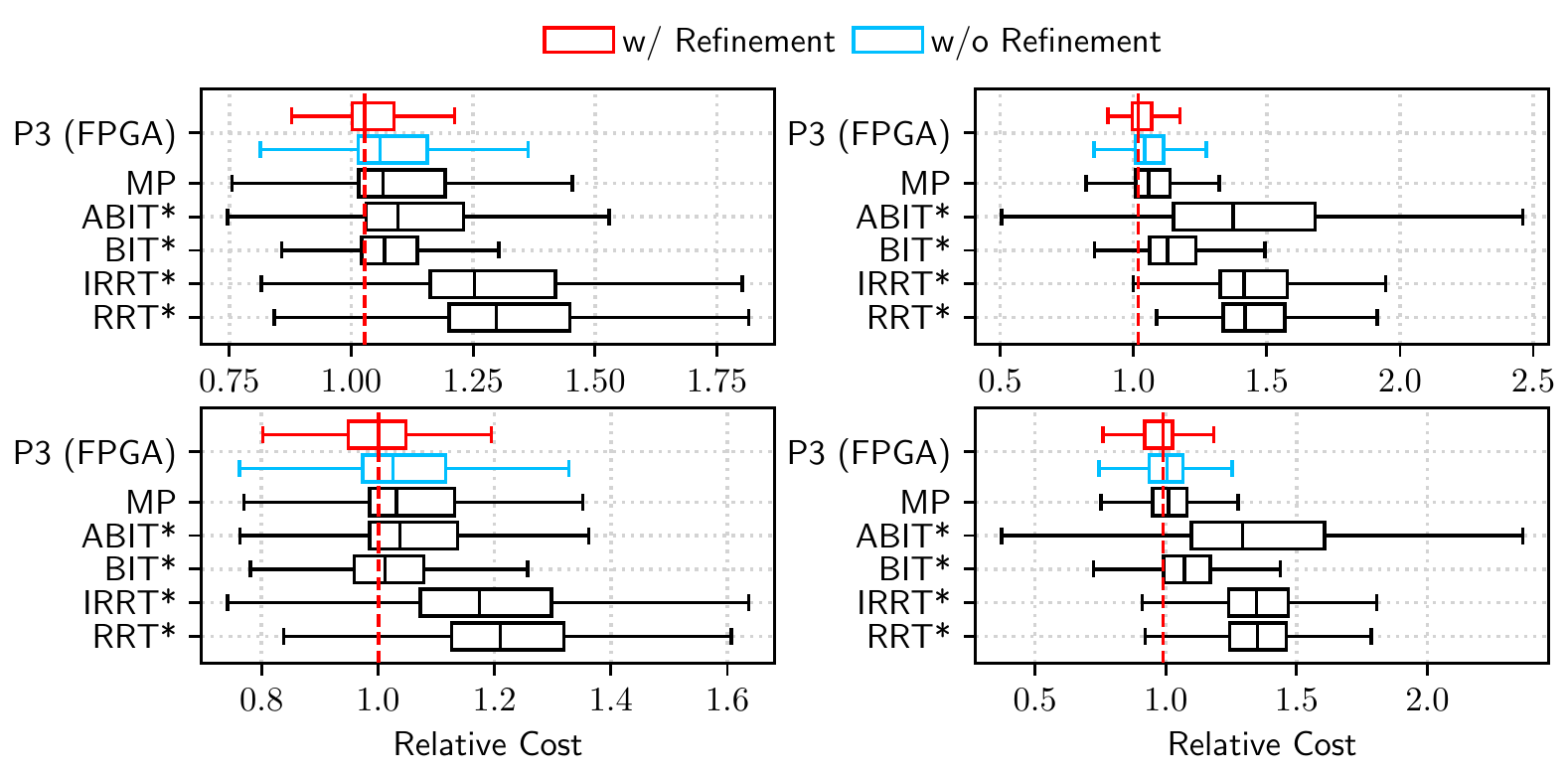}
  \caption{Distribution of the relative path cost (top: MPNet, bottom: {\ProposalName}, left: 2D, right: 3D dataset).}
  \label{fig:path-cost}
\end{figure}

This subsection evaluates the quality of solutions returned from {\ProposalName} in comparison with the other planners.
The relative path cost is used as a quality measure; it is computed by dividing a length of the output path by that of the ground-truth available in the dataset\footnote{Since ground-truth paths were obtained by running sampling-based planners with a large number of iterations, the relative cost may become less than one when a planner finds a better path with a smaller cost than the ground-truth.}.
Fig. \ref{fig:path-cost} shows the results on the MPNet/{\ProposalName} datasets ($I_\mathrm{Refine}$ is set to 5).

{\ProposalName} (\textcolor{ProcessBlue}{skyblue}) produces close-to-optimal solutions and an additional refinement step further improves the quality of outputs (\textcolor{Red}{red}).
On {\ProposalName}2D dataset, {\ProposalName} with/without refinement achieves the median cost of 1.001/1.026, which is comparable to the sampling-based methods (1.038, 1.012, 1.174, and 1.210 in ABIT*, BIT*, IRRT*, and RRT*).
If we apply smoothing and remove unnecessary waypoints from the solutions in these sampling-based methods, the median cost reduces to 1.036, 1.011, 0.996, and 1.000, respectively; while IRRT* and RRT* seem to provide better quality solutions, their success rates (85.35/86.00\%) are lower than {\ProposalName} (95.45\%).
These results confirm that {\NewPNet} samples waypoints that are in close proximity to the optimal path, and helps the algorithm to quickly find a solution by creating only a few waypoints.

\begin{figure}[h]
  \centering
  \includegraphics[keepaspectratio, width=0.7\linewidth]{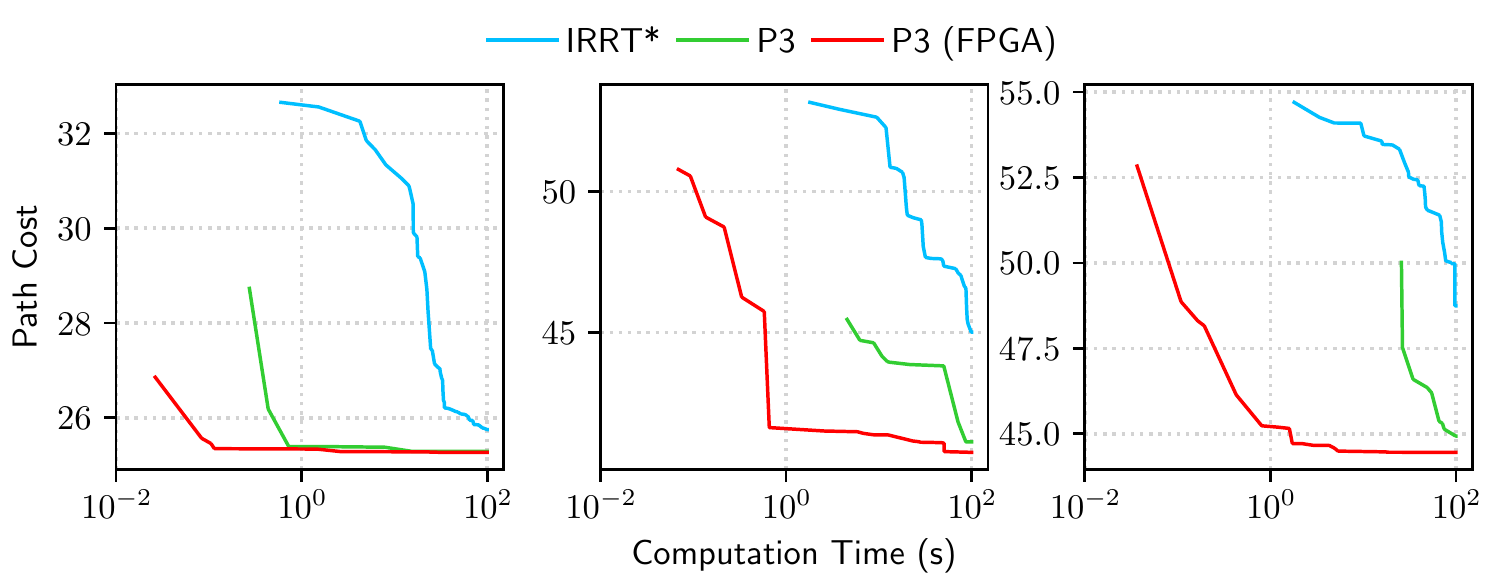}
  \caption{Evolution of the cost on {\ProposalName}2D dataset.}
  \label{fig:cost-time}
\end{figure}

Fig. \ref{fig:cost-time} plots the path costs over time on the three planning tasks taken from {\ProposalName}2D dataset.
We run {\ProposalName} and IRRT* for 100 seconds and recorded the evolution of path costs to verify the effectiveness of the refinement step.
{\ProposalNameFPGA} (\textcolor{Red}{red}) finds an initial solution within 0.1s and reaches a close-to-optimal solution within 1.0s, unlike IRRT* (\textcolor{ProcessBlue}{skyblue}) that still does not converge after 100s.
Figs. \ref{fig:result-p3net2d}-\ref{fig:result-p3net3d} show examples of the output paths obtained from MPNet and {\ProposalNameFPGA} on the {\ProposalName} dataset.

\subsection{FPGA Resource Consumption} \label{sec:eval-resource-consumption}
Table \ref{tbl:fpga-resource} summarizes the FPGA resource consumption of {\IPName} ($B = 4$).
The BRAM consumption is mainly due to the buffers for model parameters and layer outputs.
In the 2D case, {\NewEPNet} fits within the BRAM thanks to the parameter reduction and memory-efficient sequential feature extraction (Sec. \ref{sec:impl-encoder-update}), which minimizes the memory access latency and brings two orders of magnitude speedup as seen in Table \ref{tbl:inference-collision-check-time} (top left).
Since the original {\EPNet} has 32.32x more parameters, they should be stored on the external memory, which would limit the performance gain.
While {\IPName} is a package of the three modules with a collision checker and two DNNs, it consumes less than 30\% of the onboard DSP, FF, and LUT resources available in both 2D/3D cases.
These results demonstrate the resource efficiency of {\IPName}.
The batch size $B$ could be set larger (e.g., 8) for further performance (Fig. \ref{fig:mpnet-p3net-success-rate-time}).

\begin{table}[h]
  \small
  \centering
  \caption{FPGA resource consumption of {\IPName} ($B = 4$)}
  \label{tbl:fpga-resource}
  \begin{tabular}{l|r|rrrrr} \hline
    & & BRAM & URAM & DSP & FF & LUT \\ \hline
    & Total & 312 & 96 & 1728 & 460800 & 230400 \\ \hline
    \multirow{2}{*}{2D} & Use & 275 & -- & 462 & 38656 & 57404 \\
    & \% & 88.14 & -- & 26.74 & 8.39 & 24.91 \\ \hline
    \multirow{2}{*}{3D} & Use & 264 & 42 & 484 & 50653 & 69410 \\
    & \% & 84.62 & 43.75 & 28.01 & 10.99 & 30.13 \\ \hline
  \end{tabular}
\end{table}

\subsection{Power Efficiency} \label{sec:eval-power-efficiency}
Finally, we compare the power consumption of {\ProposalNameFPGA} with that of {\ProposalName} and ABIT*.
\texttt{tegrastats} is used to measure the power of Nvidia Jetson.
For the workstation, we used \texttt{s-tui}~\cite{AlexManuskin17} and \texttt{nvidia-smi} to record the power consumption of Intel Xeon W-2235 and Nvidia GeForce RTX 3090.
As shown in Fig. \ref{fig:xilinx-zcu104}, we measured the power of the entire ZCU104 board using a Texas Instruments INA219 power monitor (\textcolor{ProcessBlue}{skyblue}).
We run each planner with {\ProposalName}2D/3D datasets for five minutes and averaged the power consumption.
To exclude the power consumption of peripherals (e.g., switches and LEDs on the ZCU104), we subtracted the average power consumption in the idle state.
Table \ref{tbl:power} presents the results.

In the 2D (3D) case, {\ProposalNameFPGA} consumes 124.51x, 147.80x, and 448.47x (39.64x, 45.60x, and 145.48x) less power than ABIT*, {\ProposalName} (CPU), and {\ProposalName} (GPU) on the workstation.
It also achieves 4.40--5.38/4.58--6.10x (1.36--1.77/1.55--2.07x) power savings over ABIT*/{\ProposalName} on Nvidia Jetson.
While {\ProposalNameFPGA} consumes slightly more power than {\ProposalName} in the 3D case, this indicates that the power consumption of the {\IPName} itself is at most 0.318W ($0.809 - 0.491$).
Combined with the results from Fig. \ref{fig:mpnet-p3net-time-compare} (bottom), {\ProposalNameFPGA} offers 65.22--171.71/28.35--65.54x (5.80--17.50/4.53--19.77x) power efficiency than ABIT*/{\ProposalName} on ARM Cortex and Nvidia Jetson in the 2D (3D) case.
The power efficiency reaches 422.09x, 354.73x, and 1049.42x (44.40x, 43.78x, and 133.84x) when compared with ABIT*, {\ProposalName} (CPU), and {\ProposalName} (GPU) on the workstation.

\begin{table}[h]
  \small
  \centering
  \caption{Comparison of the power consumption (W)}
  \label{tbl:power}
  \begin{threeparttable}
    \begin{tabular}{l|rrrr|rrrr} \hline
      & \multicolumn{4}{c|}{2D} & \multicolumn{4}{c}{3D} \\ \hline
      \multirow{2}{*}{Machine} & ABIT* & \multicolumn{3}{c|}{{\ProposalName}} & ABIT* & \multicolumn{3}{c}{{\ProposalName}} \\
      & CPU & CPU & +GPU & \textbf{+IP} & CPU & CPU & +GPU & \textbf{+IP} \\ \hline
      ZCU104 & 0.480 & 0.461 & -- & 0.255 & 0.480 & 0.491 & -- & 0.809 \\
      Nano & 1.373 & -- & 1.556 & -- & 1.434 & -- & 1.678 & -- \\
      Xavier & 1.123 & -- & 1.168 & -- & 1.097 & -- & 1.250 & -- \\
      WS & 31.75 & 37.69 & 114.36\tnote{*} & -- & 32.07 & 36.89 & 117.69\tnote{*} & -- \\ \hline
    \end{tabular}
    \begin{tablenotes}
      \scriptsize
      \item[*] 114.36W: 29.72 + 84.64 (CPU/GPU); 117.69W: 32.51 + 85.17 (CPU/GPU)
    \end{tablenotes}
  \end{threeparttable}
\end{table}

\section{Conclusion} \label{sec:conc}
In this paper, we have presented a new learning-based path planning method, {\ProposalName}.
{\ProposalName} aims to address the limitations of the recently-proposed MPNet by introducing two algorithmic improvements: it (1) plans multiple paths in parallel for computational efficiency and higher success rate, and (2) iteratively refines the solution.
In addition, {\ProposalName} (3) employs hardware-amenable lightweight DNNs with 32.32/5.43x less parameters to extract robust features and sample waypoints from a promising region.
We designed {\IPName}, a custom IP core incorporating neural path planning and collision checking, to realize a planner on a resource-limited edge device that finds a path in $\sim$0.1s while consuming $\sim$1W.
{\IPName} was implemented on the Xilinx ZCU104 board and integrated to {\ProposalName}.

Evaluation results successfully demonstrated that {\ProposalName} achieves a significantly better tradeoff between computational cost and success rate than MPNet and the state-of-the-art sampling-based methods.
In the 2D (3D) case, {\ProposalNameFPGA} obtained 24.54--149.57x and 6.19--115.25x (10.03--59.47x and 3.38--28.76x) average speedups over ARM Cortex and Nvidia Jetson, respectively, and its performance was even comparable to the workstation.
{\ProposalNameFPGA} was 28.35--1049.42x and 65.22--422.09x (4.53--133.84x and 5.80--44.40x) more power efficient than {\ProposalName} and ABIT* on the Nvidia Jetson and workstation, showcasing that FPGA SoC is a promising solution for the efficient path planning.
We also confirmed that {\ProposalName} converges fast to the close-to-optimal solution in most cases.

{\ProposalName} is currently evaluated on the simulated static 2D/3D environments, and the robot is modeled as a point-mass.
As a future work, we plan to extend {\ProposalName} to more complex settings, e.g., multi-robot problems, dynamic environments, and higher state dimensions.
{\IPName} employs a standard fixed-point format for DNN inference; while it already provides satisfactory performance improvements, low-precision formats and model compression techniques (e.g., pruning, low-rank factorization) could be used to further improve the resource efficiency and speed.

\renewcommand{\baselinestretch}{1.0}
\bibliographystyle{unsrt}



\vfill

\end{document}